%% file: main.tex
\documentclass[10pt,twocolumn,letterpaper]{article}

\usepackage{cvpr}              % To produce the CAMERA-READY version
\input{preamble}

\definecolor{cvprblue}{rgb}{0.21,0.49,0.74}
\usepackage[pagebackref,breaklinks,colorlinks,allcolors=cvprblue]{hyperref}
\usepackage{xcolor}
\usepackage{multirow}
\usepackage{booktabs}
\usepackage{graphicx} % for \rotatebox
\usepackage[table]{xcolor} 
\usepackage{makecell}
\usepackage{pifont}
\usepackage{amssymb}

\makeatletter
\renewcommand*{\@fnsymbol}[1]{\ensuremath{\ifcase#1\or *\or \dagger\or \ddagger\or
    \mathsection\or \mathparagraph\or \|\or **\or \dagger\dagger
    \or \ddagger\ddagger \else\@ctrerr\fi}}
\makeatother

\title{Towards Universal Computational Aberration Correction\\in Photographic Cameras: A Comprehensive Benchmark Analysis}
\author{
Xiaolong Qian$^{1,*}$ \qquad Qi Jiang$^{1,}$\thanks{Equal contribution.} \qquad Yao Gao$^{1}$ \qquad Lei Sun$^{2,}$\thanks{Corresponding authors.} \qquad Zhonghua Yi$^{1}$\\Kailun Yang$^{3}$ \qquad Luc Van Gool$^{2}$ \qquad Kaiwei Wang$^{1,\dagger}$\\
\normalsize $^{1}$Zhejiang University \quad $^{2}$INSAIT, Sofia University ``St. Kliment Ohridski'' \quad $^{3}$Hunan University
}

\begin{document}
\maketitle

\input{sec_camera_ready/0_abstract}   
\input{sec_camera_ready/1_intro}

\input{sec_camera_ready/2_related}
\input{sec_camera_ready/3_problem}
\input{sec_camera_ready/4_benchmark}
\input{sec_camera_ready/5_exp}
\input{sec_camera_ready/6_conclusion}

\clearpage
\input{sec_camera_ready/ack}
{
    \small
    \bibliographystyle{ieeenat_fullname}
    \bibliography{main}
}

\input{supp}

\end{document}

%% file: preamble.tex
\newcommand{\ourdataset}{\textsc{UniCAC}}

\usepackage[table]{xcolor}
\usepackage{colortbl}
\definecolor{tableHeadGray}{gray}{.9}
\usepackage{multirow}
\usepackage{float}
\usepackage{adjustbox}
\usepackage{etoolbox}
\usepackage{dsfont}
\newcommand{\pub}[1]{{\color{gray!90}{\tiny{[{#1}]\!}}}}

\usepackage{tabularray}
\usepackage{arydshln}
\usepackage{bm}
\usepackage{amsmath}

\DeclareMathOperator*{\argmin}{arg\,min}
\usepackage{xcolor}

\def\eg{\textit{e.g.}}

\def\etal{\textit{et al}.}

%% file: sec_camera_ready/0_abstract.tex
\begin{abstract}
Prevalent Computational Aberration Correction (CAC) methods are typically tailored to specific optical systems, leading to poor generalization and labor-intensive re-training for new lenses. Developing CAC paradigms capable of generalizing across diverse photographic lenses offers a promising solution to these challenges. However, efforts to achieve such cross-lens universality within consumer photography are still in their early stages due to the lack of a \textit{comprehensive} benchmark that encompasses a sufficiently wide range of optical aberrations. Furthermore, it remains unclear \textit{which} specific factors influence existing CAC methods and \textit{how} these factors affect their performance. In this paper, we present comprehensive experiments and evaluations involving 24 image restoration and CAC algorithms, utilizing our newly proposed \ourdataset, a large-scale benchmark for photographic cameras constructed via automatic optical design. The Optical Degradation Evaluator (ODE) is introduced as a novel framework to objectively assess the difficulty of CAC tasks, offering credible quantification of optical aberrations and enabling reliable evaluation. Drawing on our comparative analysis, we identify three key factors -- \textit{prior utilization}, \textit{network architecture}, and \textit{training strategy} -- that most significantly influence CAC performance, and further investigate their respective effects. We believe that our benchmark, dataset, and observations contribute foundational insights to related areas and lay the groundwork for future investigations. Benchmarks, codes, and \textit{Zemax} files will be available at \url{https://github.com/XiaolongQian/UniCAC}.
\end{abstract}

%% file: sec_camera_ready/1_intro.tex
\section{Introduction}
\label{sec:intro}

Computational Aberration Correction (CAC) remains a longstanding challenge in computational imaging, serving as an image post-processing technique to address residual aberrations in optical systems. 
Unlike traditional Image Restoration (IR) methods that primarily focus on uniform degradations~\cite{liang2021swinir,chen2022simple}, CAC addresses spatially varying aberrations (\textit{e.g.}, field-dependent Point Spread Functions (PSFs) caused by lens imperfections) as well as channel disparities (\textit{e.g.}, chromatic aberrations from wavelength-dependent dispersion).
As sensor resolutions continue to increase and design constraints become more stringent within modern optical imaging systems, lens design has grown increasingly challenging, which underscores the critical role of CAC in preserving image quality.
\begin{figure}[!t]
  \centering
  \includegraphics[width=1.0\linewidth]{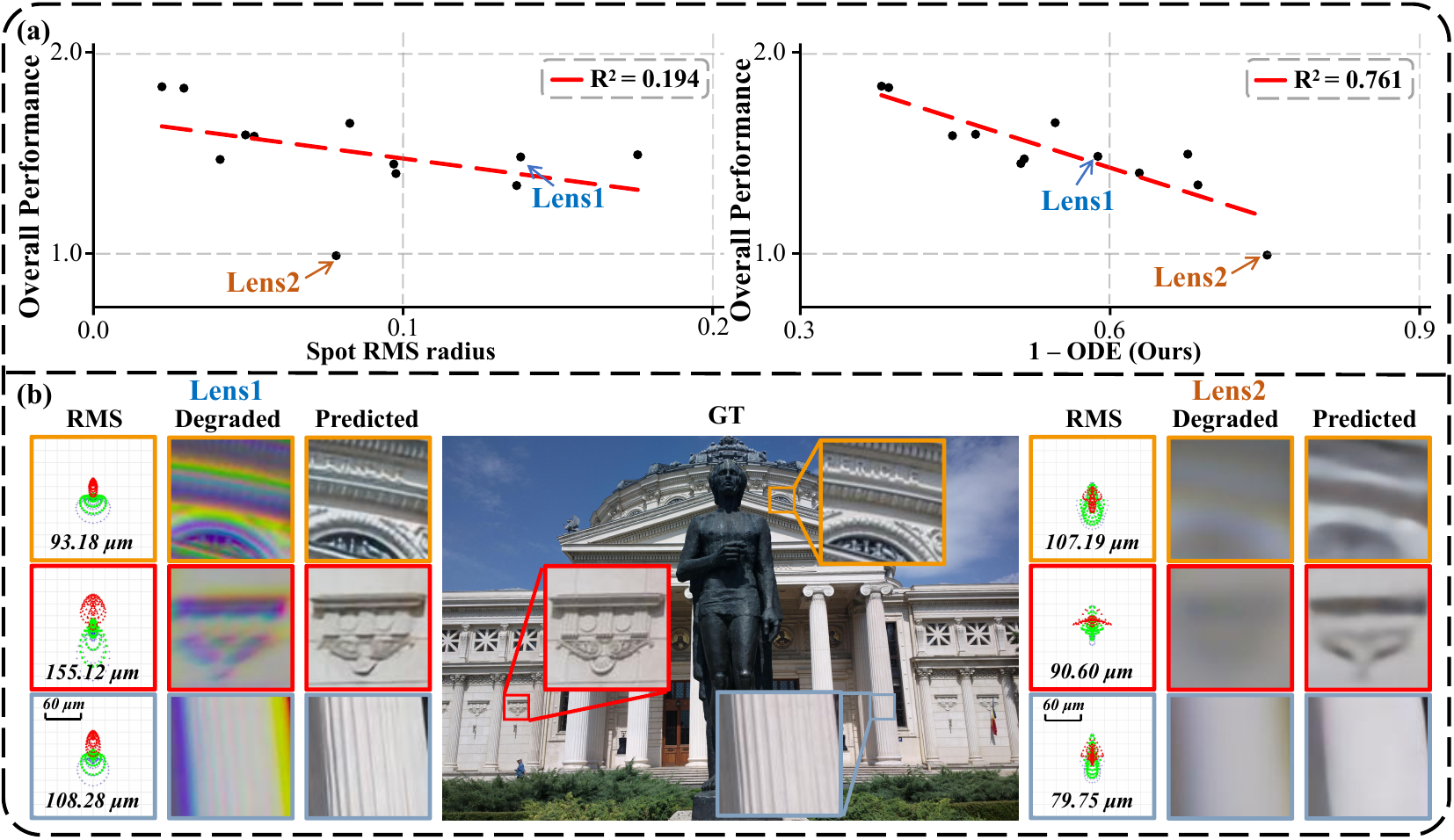}
  \vskip -0.3\baselineskip plus -1fil
  \caption{\textbf{Comparison of the linear relationship between \textit{RMS radius} and \textit{ODE} with the CAC results.} (a) We select $12$ lenses and use the same model~\cite{chen2022real} for lens-specific training. The red line shows the least squares linear fit, with $R^2$ indicating correlation strength (closer to $1$ means stronger correlation). (b) Visual results: Lens1, with a larger RMS radius, retains more details, while Lens2, with a smaller RMS radius, loses fine structures.} 
  \vskip -1.0\baselineskip plus -1fil
  \label{fig:ode analysis}
\end{figure}
By closely scrutinizing the current \textit{de facto} CAC paradigm, one can discern that they~\cite{peng2019learned,chen2021extreme_quality,jiang2023annular,qian2025towards} primarily focus on enhancing the performance of the proposed approach for specific optical systems, resulting in poor generalization and a labor-intensive retraining process for unseen lenses.

To overcome these shortcomings, \textit{universal CAC} paradigms trained on comprehensive datasets aim to generalize across unseen lenses. 
While optical aberrations are inherent to a broad spectrum of specialized instruments (\textit{e.g.}, microscopes and telescopes), our focus lies within the pervasive domain of consumer photography. 
Therefore, in this paper, we explicitly define ``universal CAC'' as the pursuit of cross-lens universality within photographic cameras.
However, the development of such universal CAC remains nascent~\cite{li2021universal,gong2024physics,jiang2024flexible}, primarily hindered by the absence of a comprehensive benchmark covering diverse photographic lens designs, largely because most commercial lens configurations are not available.
Moreover, the specific determinants that impact universal CAC and their respective effects on its performance remain unexplored, further impeding the development of this field.

Fortunately, automatic optical design, which utilizes heuristic global search algorithms~\cite{guo2019new,zhang2020automated,yue2022adaptive} or learning-based methods~\cite{cote2019extrapolating,cote2021deep,yang2023curriculum}, aims to minimize or even eliminate human intervention in the design process, and this makes it possible to obtain a large number of lens description files that conform to physical constraints.

In light of this, establishing a comprehensive universal CAC benchmark covering the potential aberration behaviors of diverse lenses is both necessary and urgent.
To this end, we extend the recent automatic optical design method OptiFusion~\cite{gao2024global} by redefining spherical parameter definitions to further include aspherical parameters.
This extension allows for designing a wide range of specifications for both spherical and aspherical lenses, thereby enhancing the versatility and applicability of our approach (see \S\ref{section:Automatic Optical Design}).

To facilitate credible benchmarking and comprehensively characterize numerous aberrations, we propose a new aberration quantization framework named \textit{Optical Degradation Evaluator} (ODE), which quantifies the difficulty of CAC and replaces the traditional RMS radius metric (see \S\ref{section:Optical Degradation Evaluation}).
This framework integrates image fidelity metrics (PSNR, SSIM~\cite{wang2004image}) with MTF-based OIQE~\cite{jiang2024minimalist} to evaluate optical characteristics.
As shown in Fig.~\ref{fig:ode analysis}, ODE exhibits higher linearity with final CAC performance compared to the RMS radius. 
The overall performance reflects our evaluation of CAC results across multiple aspects, including image fidelity, perceptual quality, and optical characteristics.

Before conducting benchmark evaluations, we first validate the reliability of our simulated aberration images from two perspectives: (1) by comparing them with both \textit{Zemax}-generated and real-captured images, and (2) by demonstrating the diversity of our benchmark across all optical degradation dimensions (see \S\ref{section:Validation of Benchmark Effectiveness}).

Building upon our constructed benchmark, we conduct a comprehensive evaluation of 24 models spanning two major categories: CAC and IR.
Through comprehensive experiments, we report nine key observations from multiple perspectives:
\textbf{Utilization of Prior:}
1) Optical Priors: both Field of View (FoV) information and PSF cues play a significant role in handling spatially varying aberrations.
2) Clear Image Priors: the codebook in FeMaSR and the stable diffusion in DiffBIR contain extensive clear image priors, which are highly beneficial for CAC.
\textbf{Architecture:} CNN-based models offer a better trade-off between CAC performance and inference time, as their convolutions efficiently capture local features and align with the nature of aberration degradation.
\textbf{Training strategy:}
Regression-based training methods enhance image fidelity; GAN-based and diffusion-based methods improve perceptual quality. 
Training paradigms for improving optical quality have been less explored. 
How to implement a comprehensive training approach to effectively boost the overall CAC performance of models is worth exploring.

In a nutshell, our contributions are threefold:

\begin{itemize}

\item 
We are the first to benchmark a wealth of image restoration models and CAC models. 
A new benchmark, named \ourdataset, is constructed by extending the recently advanced AOD method and incorporates both spherical and aspherical lenses to provide a more comprehensive representation of optical systems.
\item A new aberration quantization framework, named ODE, is proposed to quantify the difficulty of CAC, aiding in rational lens selection and enabling more credible benchmarking by addressing limitations of traditional metrics.
\item Through the experiments, we provide key insights into the factors influencing CAC, including the utilization of prior knowledge, model architecture, and training scheme.

\end{itemize}

%% file: sec_camera_ready/2_related.tex
\section{Related Work}
\label{sec:formatting}

\noindent\textbf{Computational aberration correction.}
CAC methods improve image quality by handling residual optical aberrations. 
Benefiting from the rapid development of image restoration~\cite{wang2018esrgan,liu2021swin,zhu2023denoising}, learning-based CAC methods~\cite{chen2021extreme_quality,jiang2024minimalist,luo2024correcting} have achieved more impressive results than optimization-based methods~\cite{wiener1949extrapolation,fish1995blind,schuler2012blind,2015blind}.
However, constrained by lens-specific solutions, these models exhibit limited generalization capabilities to unseen lenses.
Recently, researchers have explored universal CAC~\cite{li2021universal,gong2024physics,jiang2024flexible} by using training datasets containing a certain range of aberration distributions, but the insufficient number of spherical and aspherical optical systems limits its universality.
To this end, we construct a more comprehensive benchmark via automated design of diverse optical systems.

\noindent\textbf{Optical aberration quantification.}
Traditional methods use Zernike coefficients~\cite{dun2020learned}, RMS wavefront error~\cite{wen2021active} to quantify aberrations, but fail to link with downstream performance~\cite{ikoma2021depth,yang2023image}. 
In image quality assessment, metrics like PSNR, SSIM~\cite{wang2004image}, and LPIPS~\cite{zhang2018image} evaluate pixel-level accuracy or perceptual similarity but do not predict CAC difficulty.
To address this, we propose the Optical Degradation Evaluator (ODE) framework to quantify the CAC difficulty by integrating traditional image quality metrics with optical evaluation functions.

\noindent\textbf{Benchmarks for optical aberration robustness.} 
Hendrycks~\textit{et al.}~\cite{hendrycks2019benchmarking} assess model robustness against common corruptions.
Müller~\textit{et al.}~\cite{muller2023classification} and Jiang~\textit{et al.}~\cite{jiang2024computational} use Zernike polynomials to generate aberrations for robustness assessment.
To align with physical optics, LensCorruptions~\cite{muller2025examining} benchmarks classification and detection using 100 real lens prescriptions.
In this work, we benchmark the CAC capabilities of various models under aberrations of different severity levels.

%% file: sec_camera_ready/3_problem.tex
\section{Fundamentals and Benchmark Setup}
In this section, we first introduce the basics of Universal CAC in \S\ref{subsec:pre}.
Next, we describe the automatic optical design pipeline for generating our lens library database in \S\ref{section:Automatic Optical Design}. 
Finally, in \S\ref{subsec:data_sampling}, we propose ODE for quantifying optical degradation and use it to sample lenses, ensuring diverse and balanced aberration distributions in our benchmark.

\begin{figure}[!t]
  \centering
  \includegraphics[width=0.99\linewidth]{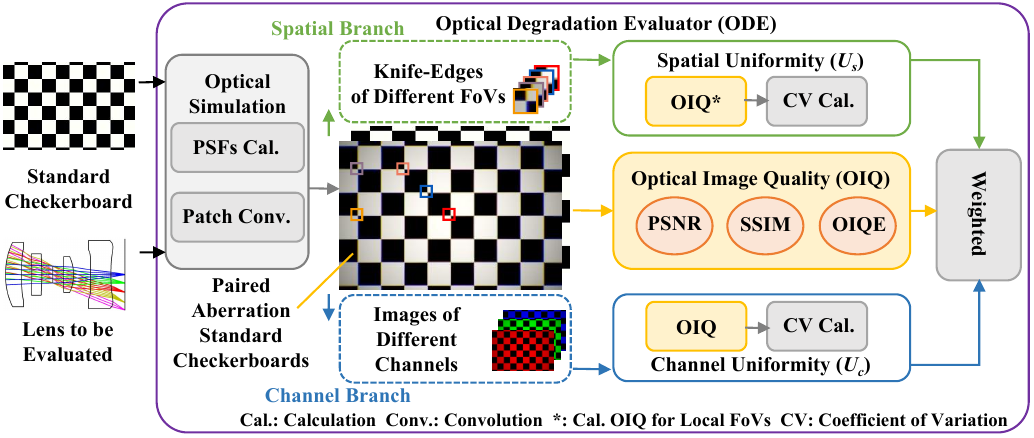}
  \vskip -0.3\baselineskip plus -1fil
  \caption{\textbf{Optical Degradation Evaluator (ODE) pipeline.} It assesses lens optical degradation severity via standard checkerboard imaging. The spatial branch calculates OIQ for various FoVs, the channel branch for different channels.} 
  \vskip -0.8\baselineskip plus -1fil
  \label{fig:ode}
\end{figure}

\subsection{Preliminaries}
\label{subsec:pre}
The objective of Universal CAC~\cite{li2021universal,jiang2024flexible} is to train a universal model $f_{\bm{\theta}}$ to restore aberration images $y_{i}$ from any unknown lens $L_{i}$ in a zero-shot manner.
Unlike traditional IR, lens aberration can be modeled as non-uniform channel-wise blur since the PSF varies both spatially and spectrally (\textit{i.e.}, across RGB channels).
Leveraging the high similarity between neighboring PSFs, we approximate this degradation process using patch-wise convolution.
For each image patch $p$ at channel c, the relationship between the aberration image patch $y{}^{p,c}$ and its latent clean patch $x{}^{p,c}$ is given by
\begin{equation}  
    % \tiny
    y_{i}^{p,c}=  x_{i}^{p,c}\otimes K_{i}^{p,c} + n,    
    \label{eq:universal cac}
\end{equation}
where $\otimes$ denotes 2D convolution, $i$ represents the selected lens from UniCACLib, and $n$ is the noise.
The universal model $f_{\bm{\theta}}$ learns a mapping from $y$ to $x$ by minimizing a loss function $L$ across the dataset:
\begin{equation}
    \bm{\theta} ^{*} =\argmin_{\bm{\theta}} \sum_{i=1}^{N}L(f_{\bm{\theta}}(y_{i}),x_{i}),
    \label{eq:loss function}
\end{equation}
where $N$ denotes the total number of lenses in UniCACLib. 
This universal model is anticipated to address the optical degradation of any unseen optical lens. 
To this end, a benchmark encompassing a wide range of diverse and realistic optical degradation is crucial for model evaluation. 
\subsection{Lens Generation by Automatic Optical Design}
\label{section:Automatic Optical Design}
Compared to synthetic optical degradation generated using random Zernike models~\cite{muller2023classification, jiang2024computational}, degradation caused by optimized, manufacturable lenses better reflects real-world performance and thus provides a more robust assessment of models in practical CAC scenarios~\cite{jiang2024flexible}.
Given the difficulty of obtaining such lenses and the labor of manual design, we propose using automated, physically-constrained optical design to generate large-scale, high-quality lens samples as the basis for our benchmark.

We extend OptiFusion~\cite{gao2024global} to automatically design a large number of both spherical and aspherical lenses, a state-of-the-art method capable of rapidly and automatically designing optical lenses of varying specifications. 
To ensure diversity in aberration characteristics, four types of specifications that deliver large impacts on the final aberration behavior, \textit{i.e.}, the piece number, aperture position, half FoV, and F number of the lens, are considered for generating diverse lens samples (detailed in the supplementary material).

\begin{figure*}[!t]
  \centering
  \includegraphics[width=0.99\linewidth]{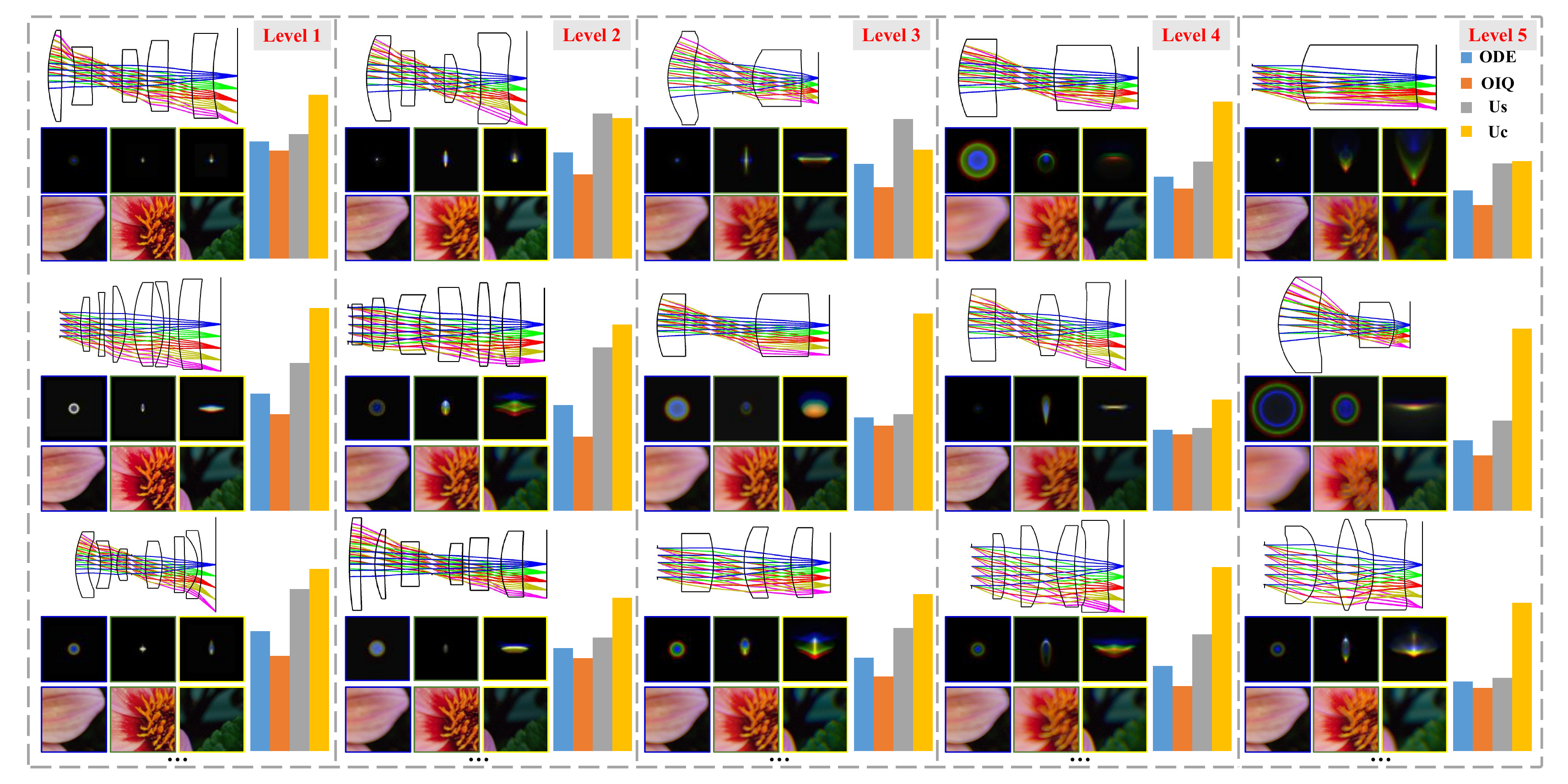}
  \vskip -0.5\baselineskip plus -1fil
  \caption{\textbf{Benchmark overview.} The sampled lenses are divided into $5$ levels based on ODE, with $3$ lenses from each level showcased to facilitate presentation. Each lens is illustrated with the ray tracing diagram, PSFs for $3$ FoVs, and corresponding aberration image patches. The distribution of quantification dimensions for each lens is also presented in a bar chart.}
  \vskip -1.0\baselineskip plus -1fil
  \label{fig:bench show}
\end{figure*}

\subsection{Lens Sampling}
\label{subsec:data_sampling}
To construct a convincing benchmark for the comprehensive evaluation of models, we sample from the constructed lens library while ensuring that the optical degradation levels span uniformly from mild to severe. 
However, as aforementioned, optical degradation is inherently multi-dimensional and lacks a unified quantification metric for sampling and classification. 
To address this, we propose the Optical Degradation Evaluator (ODE), which quantifies lens degradation based on the imaging result of a standard checkerboard captured by the lens.

\noindent\textbf{Optical degradation evaluator.}
\label{section:Optical Degradation Evaluation}
According to previous work~\cite{zhou2024revealing}, the overall image quality of the degraded image, the severity of spatial variation, and the chromatic aberration characteristics of optical degradation significantly influence the model's CAC performance.
Hence, the proposed ODE (shown in Fig.~\ref{fig:ode}) comprehensively assesses optical degradation by integrating three aspects: 1) Optical Image Quality ($OIQ$), which evaluates the overall fidelity and optical characteristics; 2) spatial uniformity ($U_s$), which assesses the severity of spatial variation; and 3) channel uniformity ($U_c$), which evaluates the chromatic aberration.
The calculation of ODE is formulated as:

\begin{equation}
ODE = \lambda_{oiq}~OIQ + \lambda_s~U_s + \lambda_c~U_c,
\label{eq:ODE}
\end{equation}

Specifically, OIQ incorporates traditional image quality evaluators (such as PSNR and SSIM~\cite{wang2004image}) as well as MTF-based OIQE~\cite{jiang2024minimalist} to provide an image quality assessment considering optical properties:
\begin{equation}
    \small
    OIQ = \alpha~\frac{PSNR}{50}  + \beta~\frac{SSIM - 0.5}{0.5} + \gamma~OIQE.
    \label{eq:Score}
\end{equation}
Additionally, to measure spatial uniformity ($U_s$) and chromatic aberration ($U_c$), both metrics utilize the coefficient of variation (CV) of OIQ values, which is defined as:
\begin{equation}
    U_{{s,c}}=e^{-\sigma CV_{{s,c}}},
    \label{eq:uniformity fov}
\end{equation}
where $U_s$ is calculated from the CV of OIQ values across $5$ FoVs, and $U_c$ is derived from the CV of OIQ values across $3$ channels; $\sigma$ denotes a scaling factor.

In summary, ODE serves as a quantitative measure of the overall severity of a lens's optical degradation, providing the sampling basis for constructing a foundational benchmark. 
Additionally, the $U_s$ and $U_c$ within the ODE framework assess optical degradation from the critical dimensions of spatial variation and chromatic aberration, respectively. 
Sub-benchmarks centered around these dimensions are shown in Sec.~\ref{sec:us benchmark} and Sec.~\ref{sec:uc benchmark}.

\noindent\textbf{Benchmark construction.}
Based on the proposed ODE, we sample lenses from the lens library to construct our UniCACLib, ensuring a reasonable distribution of $ODE$, $OIQ$, $U_s$, and $U_c$ to facilitate benchmarking universal CAC models from the perspectives of overall severity, spatial variation severity, and chromatic aberration characteristic of optical degradation.
In our main benchmark setting (Fig.~\ref{fig:bench show}), we divide the sampled lens data into $5$ levels based on the ODE, showing $3$ lenses from each level.
For each lens, we display the ray tracing diagram, the PSF for $3$ FoVs, and the corresponding aberration images for each field. 
Additionally, the bar chart visually presents the distribution of different quantification dimensions for each lens.

%% file: sec_camera_ready/4_benchmark.tex
\section{Benchmark Validation}
\label{section:Validation of Benchmark Effectiveness}
In this section, we validate our {\ourdataset} benchmark from two aspects: 1) the simulated aberration images are comparable to real-world ones in \S\ref{sec:benchmark_val}, and 2) our benchmark exhibits diversity across all quantification dimensions of optical degradation, providing a comprehensive evaluation of a universal CAC model's generalization capability in \S\ref{subsec:Distributions of Lens Samples}. 

\subsection{Evaluation of Aberration Simulation}
\label{sec:benchmark_val}
Given the impracticality of manufacturing numerous lenses from UniCACLib to capture test images, the test aberration images in the constructed benchmark are all generated through aberration simulation.
To ensure that the simulated data closely approximates real-world data, we employ a state-of-the-art optical simulation model~\cite{chen2021optical} and validate its reliability through the following two aspects.

\begin{figure}[!t]
  \centering
  \includegraphics[width=0.99\linewidth]{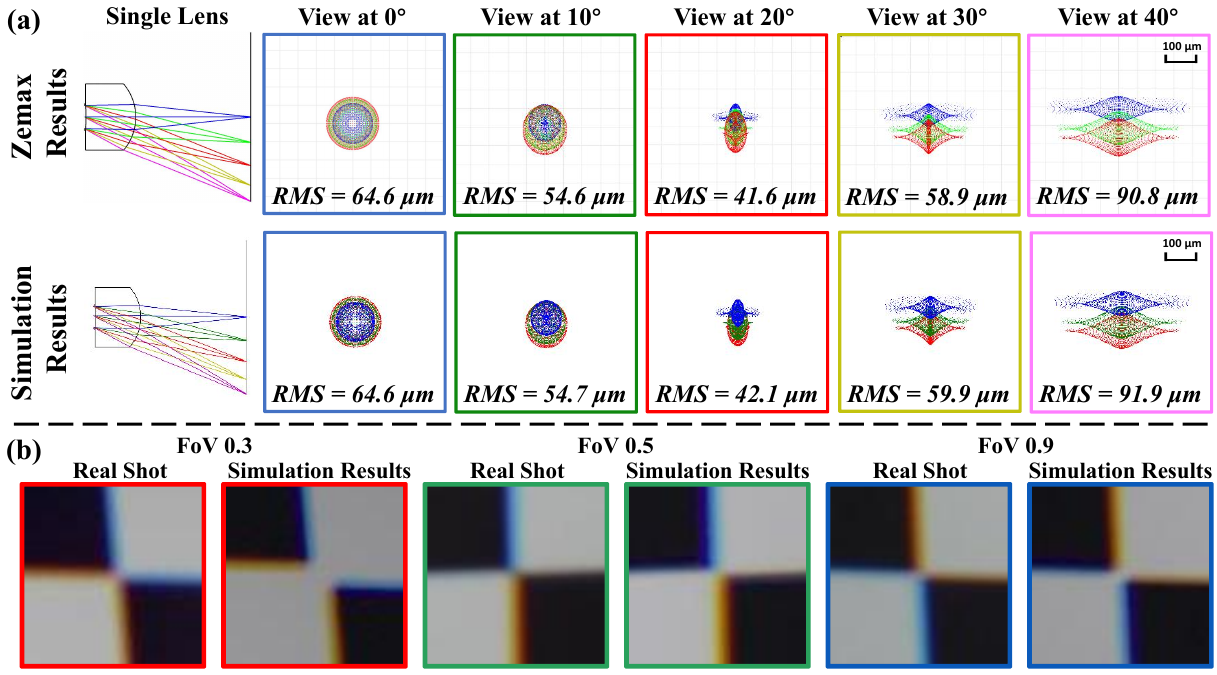}
  \vskip -0.3\baselineskip plus -1fil
  \caption{(a) The ray tracing diagram, spot diagrams and RMS spot sizes produced by the simulation framework closely resemble those obtained from \textit{Zemax}. The tested system is a single-lens optical system. (b) Comparison between simulation and real-shot aberration images.} 
  \label{fig:compare ZMX RMS}
  \vskip -0.5\baselineskip plus -1fil
\end{figure}

\noindent\textbf{Comparison with \textit{Zemax}.}
We compare the simulation results with those from \textit{Zemax} using a single-lens optical system for this analysis. 
The ray tracing outcomes from both \textit{Zemax} and the used system are visualized, where the RMS spot sizes are subsequently calculated. 
As illustrated in Fig.~\ref{fig:compare ZMX RMS}(a), the spot sizes and ray distribution exhibit nearly identical patterns across all fields of view, with an average error margin of only $1{\mu}m$. 
This indicates that the optical simulation process is precise enough to produce PSFs.

\noindent\textbf{Simulation \textit{vs.} Real shot.}
We further validate the fidelity of the simulation system by comparing simulated results with real-captured images.
As shown in Fig.\ref{fig:compare ZMX RMS}(b), across different FoVs, the simulated images exhibit aberration characteristics (\eg, color fringes) similar to those in real-world captures.
The remaining discrepancies mainly stem from errors in generating the pseudo-GT: after capturing real aberration images, we apply edge extraction and colorization to obtain a pseudo-GT~\cite{chen2021extreme_quality}(detailed in the supplementary material).

Overall, the applied simulation model is capable of producing aberration images comparable to those captured with the corresponding real-world lens, serving as an effective method for generating benchmark test images.

\begin{figure}[!t]
  \centering
  \includegraphics[width=0.99\linewidth]{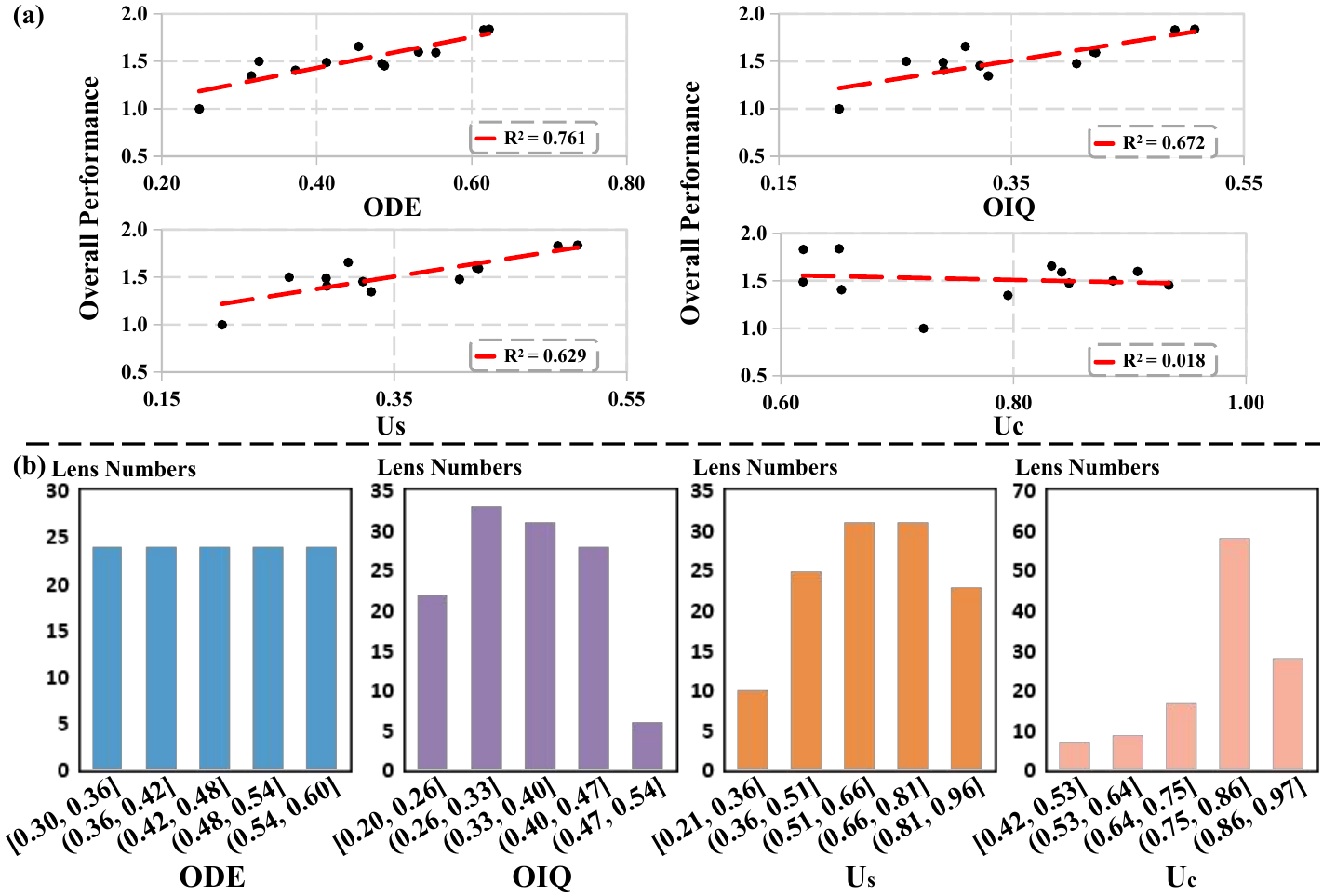}
  \vskip -0.5\baselineskip plus -1fil
  \caption{(a) Relevance between ODE components and CAC performance. (b) Distribution of different quantification dimensions in our sampled lens library.} 
  \vskip -1.0\baselineskip plus -1fil
  \label{fig:sampled lens lib distribution}
\end{figure}
\subsection{Lens Sample Distributions}
\label{subsec:Distributions of Lens Samples}
Similar to Fig.\ref{fig:ode analysis}, we perform lens-specific training and testing on $12$ lenses sampled from our library to examine the influence of each ODE component on CAC performance.
As shown in Fig.\ref{fig:sampled lens lib distribution}(a), results indicate that $OIQ$ and $U_s$ exhibit stronger correlations with final CAC outcomes, suggesting they play a more significant role, while $U_c$ correlates weakly.
Given the importance of chromatic aberration correction in CAC, we retain $U_c$ in ODE but assign it a lower weight (see Sec.~\ref{sec:coefficient selection}).
ODE, as a combination of the three components, shows the highest linear relationship with the CAC results, demonstrating its effectiveness in quantifying optical degradation.
These evidences support our lens sampling based on ODE to construct our benchmark.
Fig.~\ref{fig:sampled lens lib distribution}(b) presents the distribution of the 120 sampled lenses in UniCACLib across different quantification dimensions.
For each dimension, our UniCACLib contains samples with values ranging from low to high, supporting benchmark evaluations of methods in the relevant dimensions. 
Our sampling primarily ensures a uniform distribution along the main benchmark dimension, \textit{i.e.}, the ODE, but the samples in other dimensions are also sufficient to form sub-benchmarks.
These evidences illustrate our benchmark's diversity across all quantification dimensions of optical degradation, which can provide a comprehensive evaluation of a universal CAC model’s generalization capability.
To further validate the rationality of our benchmark, we compare its aberration distribution with representative ZeBase lenses via t-SNE, showing broad coverage of common aberrations (see supplementary material for more details).

\subsection{Evaluation Metric}
To comprehensively evaluate the CAC performance of each method, we employ six metrics categorized into three dimensions: image fidelity (PSNR, SSIM~\cite{wang2004image}), optical quality (OIQE~\cite{jiang2024minimalist}), and perceptual quality (LPIPS~\cite{zhang2018image}, FID~\cite{heusel2017gans}, ClipIQA~\cite{wang2022exploring}). 
Referencing NTIRE2024-RAIM~\cite{liang2024ntire}, we introduce an Overall Performance (O. P.) for comprehensive evaluation, with its coefficients set as follows to ensure balanced contributions from these three aspects (see Sec.~\ref{sec:coefficient selection}):
\vskip -0.5\baselineskip
{\footnotesize
\begin{equation}
    \begin{aligned}
        O. P. &= 0.4 \times \frac{PSNR}{50} + 0.3 \times \frac{SSIM - 0.5}{0.5} + 0.4 \times \frac{1 - LPIPS}{0.4} \\
              &\quad + 0.3 \times OIQE + 0.1 \times \frac{100 - FID}{100} + 0.1 \times ClipIQA.
        \label{eq:OP}
    \end{aligned}
\end{equation}
}
\vskip -0.5\baselineskip

\begin{figure}[!t]
  \centering
  \includegraphics[width=0.99\linewidth]{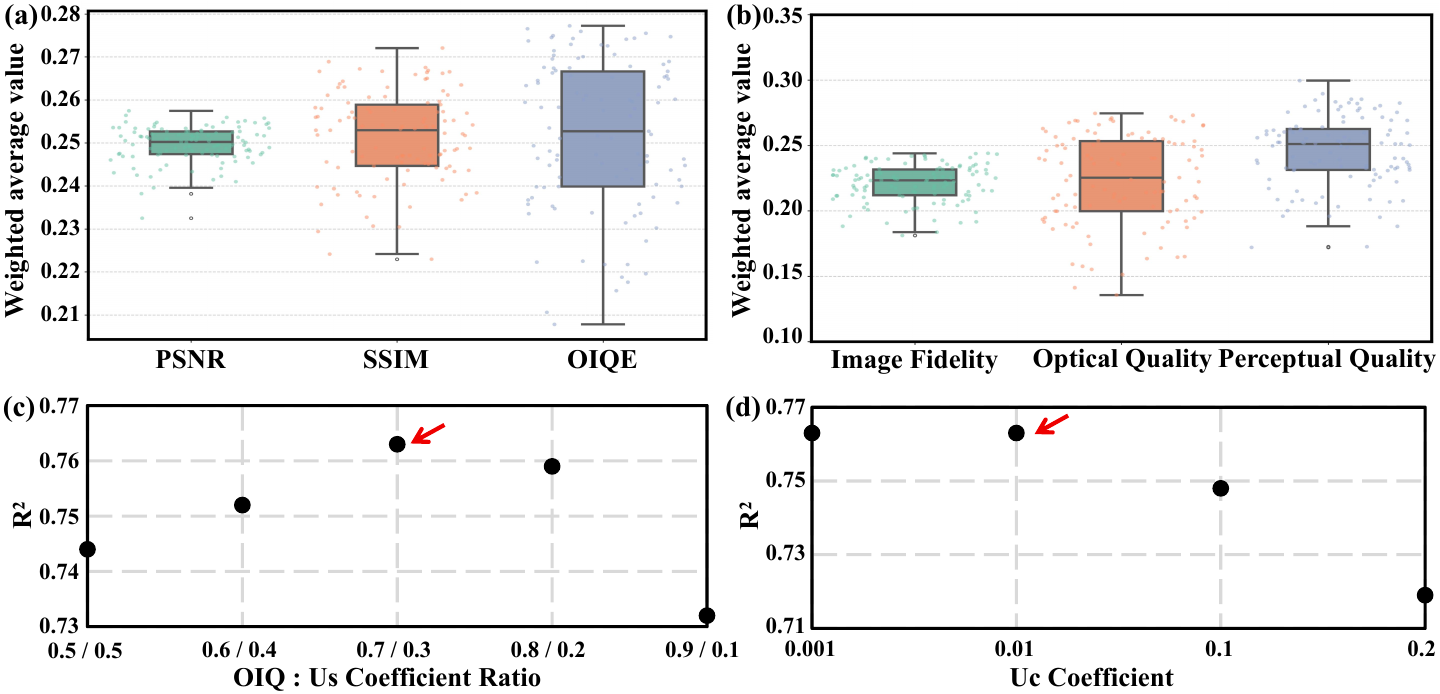}
  \vskip -0.8\baselineskip plus -1fil
  \caption{(a)~Weighted average value of PSNR, SSIM, and OIQE. (b)~Weighted average value of Image Fidelity, Optical Quality, and Perceptual Quality. (c)~Effect of varying coefficient ratios of $OIQ$ and $U_s$ on the linearity of ODE and O. P. (d)~Effect of varying coefficient ratios of $U_c$ on the linearity of ODE and O.~P.}
  \vskip -0.5\baselineskip plus -1fil
  \label{fig:re_coe}
\end{figure}
\subsection{Coefficient Selection Rationale}
\label{sec:coefficient selection}
\noindent\textbf{Weight coefficient.}
we set $\lambda_{oiq} = 0.7$, $\lambda_s = 0.3 $, and $\lambda_c = 0.01$, $\alpha = 0.4$, $\beta = 0.3$, $\gamma = 0.3$, $\sigma =5$, with an analysis of these choices provided below.

\noindent\textbf{OIQ:}
Following the 4:3 PSNR–SSIM weighting scheme adopted in NTIRE2024-RAIM~\cite{liang2024ntire}, we extend the evaluation by incorporating the OIQE metric, which quantifies optical image quality on a [0, 1] scale.
To maintain the original PSNR–SSIM balance while integrating optical fidelity, we set the final weights to 4:3:3.
As shown in Fig.~\ref{fig:re_coe}(a), we report the weighted averages of PSNR, SSIM, and OIQE computed from the reconstructed outputs of the model~\cite{cho2021rethinking} on the test lens.
This analysis confirms that all three metrics contribute comparably to the overall evaluation.

\noindent\textbf{O. P.}:
Following NTIRE2024-RAIM~\cite{liang2024ntire} and the OIQ metric, we assign weights of 4:3:4:3 to PSNR, SSIM, LPIPS, and OIQE, respectively.
% For FID and ClipIQA, the coefficients are chosen to ensure a balanced contribution across the three evaluation dimensions in the Overall Performance (O.P.): image fidelity (PSNR, SSIM), optical quality (OIQE), and perceptual quality (LPIPS, FID, ClipIQA).
For FID and ClipIQA, coefficients are chosen to balance the three dimensions of Overall Performance (O.P.): image fidelity (PSNR, SSIM), optical quality (OIQE), and perceptual quality (LPIPS, FID, ClipIQA).
Accordingly, we set their relative weight to 1:1.
As illustrated in Fig.~\ref{fig:re_coe}(b), this configuration produces comparable weighted averages across the three dimensions on reconstructions from the model~\cite{cho2021rethinking}, validating the effectiveness of our balanced weighting scheme.
This design enables a holistic assessment of model performance under varying levels of optical degradation.

\noindent\textbf{ODE}:
Although chromatic aberration has a minor impact on overall linearity, we assign it a small fixed weight to reflect its significance in the CAC field.
As shown in Fig.~\ref{fig:re_coe}(c), a 7:3 weighting ratio between $OIQ$ and $U_s$ yields the highest correlation between the Optical Degradation Evaluator (ODE) and Overall Performance (O.P.).
After establishing this ratio, we further analyze the influence of $U_c$’s coefficient.
Fig.~\ref{fig:re_coe}(d) shows that setting the coefficient of $U_c$ to 0.01 maximizes the linearity between ODE and O.P.
These results are based on experiments conducted across 12 datasets, each corresponding to a different lens, where lens-specific training was performed using the same model~\cite{chen2022real}.

%% file: sec_camera_ready/5_exp.tex
\section{Experiments and Analysis}

In this section, we evaluate the CAC capabilities of $24$ methods from multiple perspectives, conducting a comprehensive benchmark that includes aspects such as the aberration severity level, image uniformity, and chromatic aberration.

\begin{table}[!t]
\captionsetup{font=normal}
\centering

\input{Table/method_all}
\vskip -0.5\baselineskip plus -1fil
\caption{Overview of the $24$ evaluated methods.}
\vskip -0.8\baselineskip plus -1fil
\label{tab:method_all}
\end{table}

\begin{table*}[!t]
\captionsetup{font=normal}
\centering

\input{Table/evaluate_all_revised}
\vskip -0.5\baselineskip plus -1fil
\caption{\textbf{Quantitative performance of methods across $5$ aberration severity levels.} O.~P. denotes overall performance. The ranking in parentheses represents the overall performance ranking of the methods. The number of parameters, FLOPs, and inference time of each method are also listed. The image with a resolution of $256{\times}256$ is used to calculate FLOPs and inference time. The best and second best performances are in \textbf{bold} and \underline{underline}.}
\vskip -1.0\baselineskip plus -1fil
\label{tab:evaluate_all}
\end{table*}

\subsection{Implementation Details}
\label{subsec:detail}
\noindent\textbf{Datasets.}
Our benchmark datasets are generated by convolving GT images with PSFs from our lens library.
The training set uses about $3,000$ GT images (Flickr2K~\cite{timofte2017ntire}, DIV2K) degraded by PSFs randomly sampled from our $873$ training lenses.
The test set uses $26$ self-captured high-resolution GT images and is convolved with PSFs from $120$ test lenses.
See the supplement for additional details.

\noindent\textbf{Model training.}
We comprehensively evaluate 24 restoration methods for CAC, detailed in Tab.~\ref{tab:method_all}, categorizing them into CAC and IR, further distinguished by blind/non-blind types and optimization/learning-based approaches within each category.
To maximize the capability of learning-based approaches, we use the official training configurations provided by different methods to train models.
To ensure a fair comparison, all models are trained without any special tricks (\textit{e.g.}, TLC in NAFNet and EMA in SwinIR).

Additionally, different methods may not use exactly the same metric calculations in their papers. 
Therefore, we use the popular open-source toolbox PyIQA~\cite{pyiqa} to calculate metrics, except for FID and OIQE.
For FID and OIQE, we rely on the official implementations~\cite{Seitzer2020FID,jiang2024minimalist}.
All the experiments of our work are implemented on A800 GPUs.

\subsection{Results for Overall CAC Performance}

To thoroughly assess the CAC performance of all $24$ methods, we present their average results across five aberration severity levels in Tab.~\ref{tab:evaluate_all}, where the levels are defined based on the proposed ODE.

\begin{figure}[!t]
  \centering
  \includegraphics[width=1.0\linewidth]{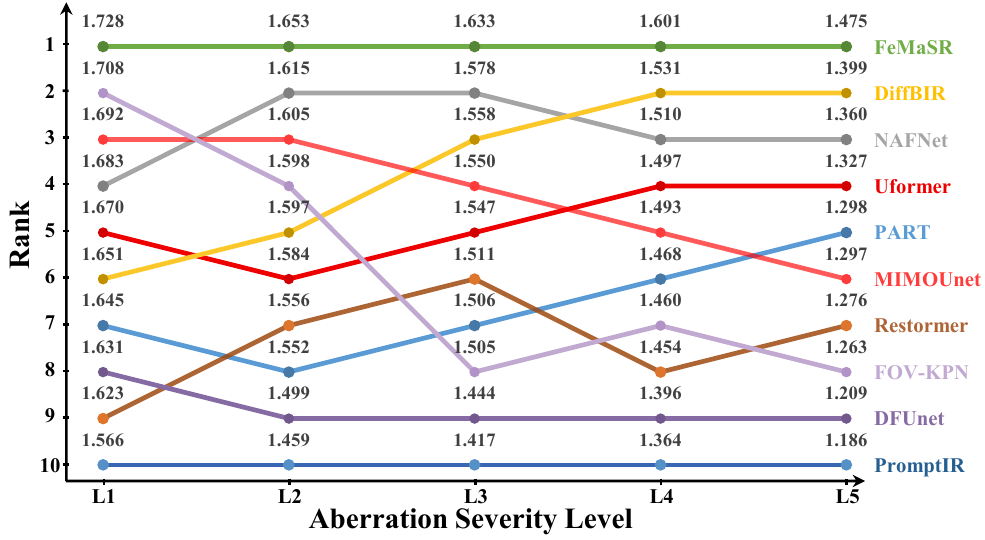}
  \vskip -0.8\baselineskip plus -1fil
  \caption{Ranking trends of different methods across aberration severity levels, with severity increasing from L1 to L5.}
  \vskip -1.0\baselineskip plus -1fil
  \label{fig:level compare}
\end{figure}

\begin{figure*}[t]
  \centering
  \includegraphics[width=0.99\linewidth]{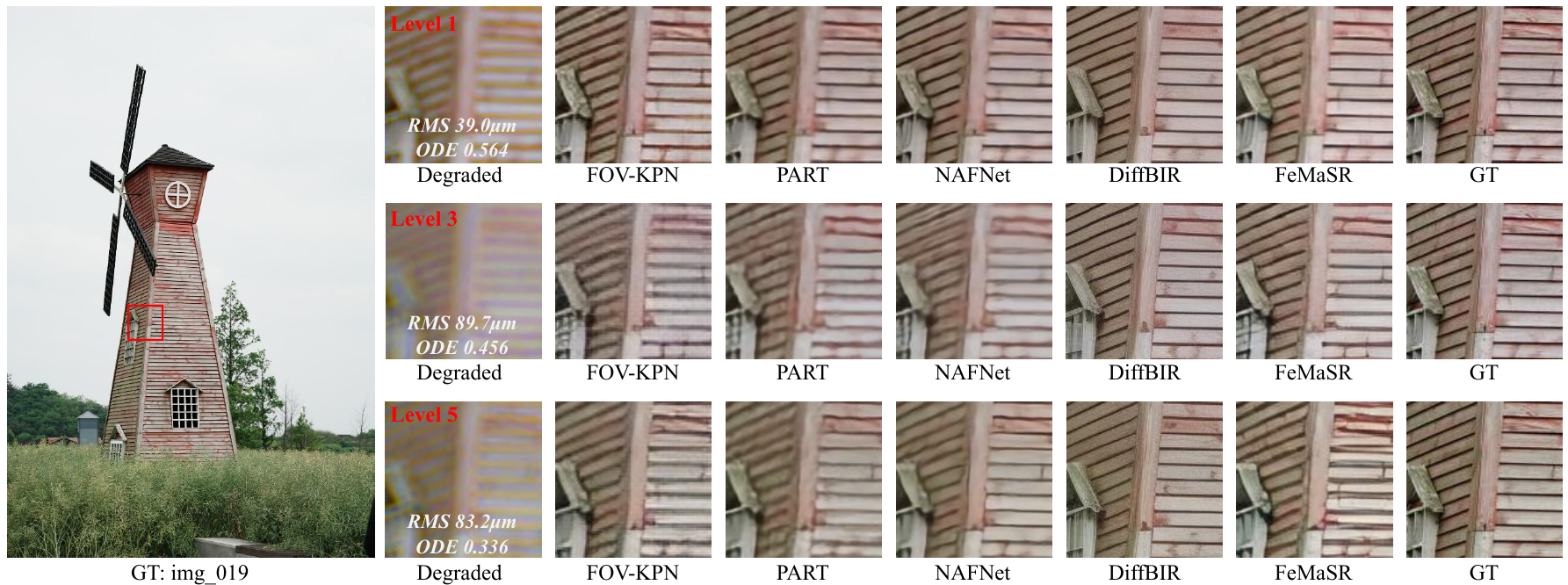}
  \vskip -0.5\baselineskip plus -1fil
  \caption{\textbf{Visual comparison of five representative methods at $3$ levels of aberration severity.} We provide RMS/ODE values to facilitate a better understanding of the relationship between quantitative values and qualitative outcomes.} 
  \vskip -1.0\baselineskip plus -1fil
  \label{fig:level compare visual results}
\end{figure*}
\noindent\textbf{Observation 1:} \textit{Learning-based methods outperform optimization-based methods.} 
Driven by data, learning-based methods exhibit excellent CAC performance and strong generalization capabilities. In contrast, optimization-based methods not only require calibration or estimation of the PSF but are also highly sensitive to noise and nonlinear factors, leading to suboptimal CAC results.

\noindent\textbf{Observation 2:} \textit{Training paradigms greatly influence CAC performance.}
Regression-based training methods enhance image fidelity, with the PART achieving the best PSNR. 
In contrast, GAN and diffusion-based methods excel in perceptual quality, with FeMaSR and DiffBIR outperforming regression methods in LPIPS and ClipIQA. 
Training strategies targeting OIQE remain underexplored and need further study for specific applications.

\noindent\textbf{Observation 3:} \textit{Proper use of PSF improves generalization in universal CAC.}
While PART and SwinIR share the same backbone network, the PSF-attention mechanism in PART effectively modulates the PSF, leading to consistently superior performance across all metrics.

\subsection{Results for Severity Level}

To explore the ability of different methods to handle varying degrees of aberrations, we select the top-$7$ IR methods and the top-$3$ CAC methods based on their overall performance for comparison.
The ranking trends of the ten selected methods are shown in Fig.~\ref{fig:level compare}. 
Furthermore, to offer a more intuitive presentation of the CAC results, visualization results are provided in Fig~\ref{fig:level compare visual results}.
Both quantitative and qualitative results demonstrate that the performance of each model declines as the degree of aberration increases.

\noindent\textbf{Observation 4:} \textit{Clear image priors are crucial for CAC.}
FeMaSR benefits from a pre-trained codebook that helps restore fine details, while DiffBIR leverages generative priors from Stable Diffusion~\cite{rombach2022high} to enhance perceptual quality via gradual denoising.

\noindent\textbf{Observation 5:} \textit{Diffusion-based methods are effective for severe aberrations.}
Methods like DiffBIR perform well under severe aberrations by using strong generative priors to synthesize realistic details.
As shown in Fig.\ref{fig:level compare}, its ranking improves with increasing aberration severity.
Fig.\ref{fig:level compare visual results} further illustrates its ability to generate plausible structures, though noise becomes more noticeable at higher degradation levels.

\noindent\textbf{Observation 6:} \textit{The effectiveness of optical priors depends on their representation capacity.}
FOV-KPN excels under mild aberrations by using low-dimensional FOV coordinates to adaptively handle local variations. 
As aberrations grow more complex, these low-dimensional priors become insufficient, while PSF-based methods like PART gain an advantage by leveraging dense 2D PSFs.
This trend is reflected in Fig.~\ref{fig:level compare}, where FOV-KPN's ranking declines as severity increases, whereas PART rises.

\noindent\textbf{Observation 7:} \textit{CNNs balance CAC performance and speed effectively.}
Regression-based CNNs (\textit{e.g.}, NAFNet, MIMOUNet) deliver high CAC results with low latency, likely because aberration degradation, inherently a convolution process, benefits from CNNs’ efficient capture of local features and spatial relationships.

\subsection{Results for Image Spatial Uniformity Level}
\label{sec:us benchmark}
\begin{figure}[!t]
  \centering
  \includegraphics[width=0.99\linewidth]{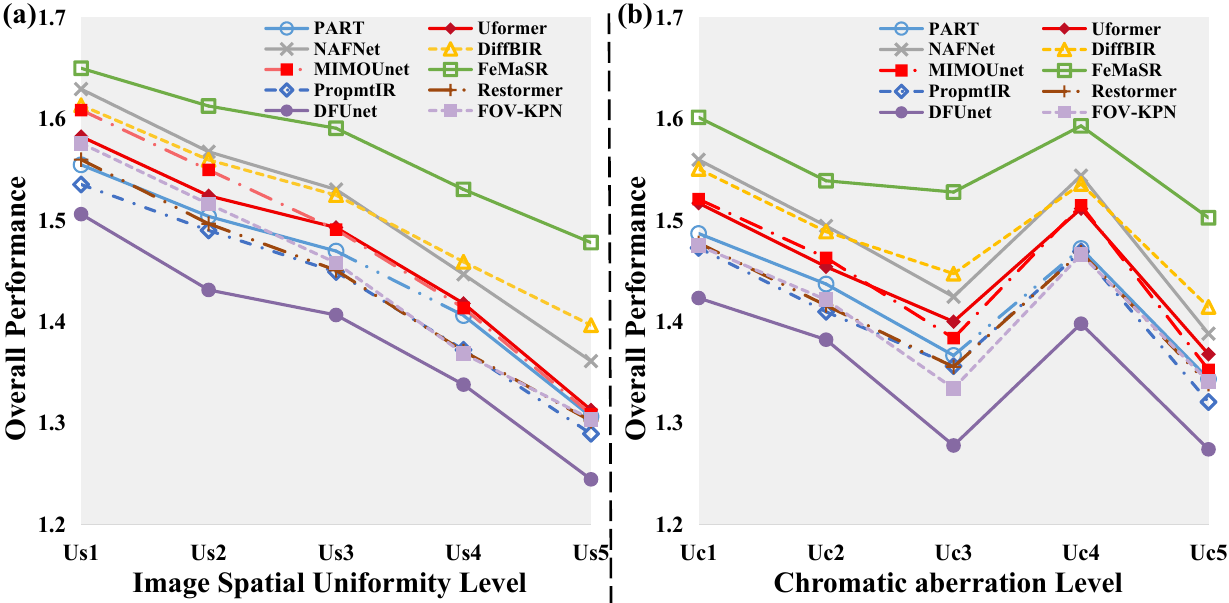}
  \vskip -0.8\baselineskip plus -1fil
  \caption{Line charts of different methods under five levels of spatial uniformity ($U_{s}$) and five levels of chromatic aberration severity~($U_{c}$).} 
  \vskip -0.5\baselineskip plus -1fil
  \label{fig:UF UC compare}
\end{figure}

Optical aberration is spatially variant, an important characteristic worthy of further exploration. 
Thus, we divide UniCACLib into 5 levels by spatial uniformity ($U_{s1}$ to $U_{s5}$), ensuring a reasonable distribution of $OIQ$ and $U_{c}$.

\noindent\textbf{Observation 8:} \textit{Image spatial uniformity significantly affects CAC performance.}
In Fig~\ref{fig:UF UC compare} (a), optical aberrations introduce spatial non-uniformity, leading to a decline in CAC performance as uniformity worsens.

\subsection{Results for Chromatic Aberrations Level}
\label{sec:uc benchmark}
Consistent with the previous sections, we divide UniCACLib into $5$ levels based on $U_c$, ensuring a reasonable distribution of $OIQ$ and $U_s$ within each level.

\noindent\textbf{Observation 9:} \textit{Within refractive photographic lenses, chromatic aberration has limited impact on CAC performance.}
Fig.~\ref{fig:UF UC compare}(b) shows CAC performance lacks clear correlation with increasing chromatic aberration severity.

%% file: Table/method_all.tex
\resizebox{\columnwidth}{!}{%
\setlength\tabcolsep{3pt} % 设置表格列之间的间距
\renewcommand\arraystretch{1.02} % 设置表格行之间的间距  
\begin{tabular}{crccc}
% \hline
\Xhline{2\arrayrulewidth} % 底部加粗横线（宽度为默认的2倍）
\rowcolor{tableHeadGray}
\multicolumn{2}{c}{Model}            & Blind? & Model Type             & Learning Paradigm  \\ \hline
\multirow{10}{*}{CAC} & WienerDeconv\pub{MITPress49}~\cite{wiener1949extrapolation} & \ding{55}      & Optimization     & -                                                                                                                       \\
                      & FSANet\pub{OE22}~\cite{lin2022non}       & \ding{55}      & Optimization\&CNN & Regression                                                                                                               \\
                      & PART\pub{TIP24}~\cite{jiang2024minimalist}         & \ding{55}      & Transformer      & Regression                                                                                                               \\ \arrayrulecolor{gray}\cdashline{2-5}\arrayrulecolor{black}
                      
                      & Peng ~\etal\pub{TOG19}~\cite{peng2019learned}   & \ding{51}      & CNN              & GAN                                                                                               \\
                      & DFUnet\pub{TOG21}~\cite{chen2021optical}       & \ding{51}      & CNN              & Regression                                                                                                            \\
                      & FOV-KPN\pub{ICCV21}~\cite{chen2021extreme_quality}       & \ding{51}      & CNN              & Regression                                                                                                               \\
                      & Eboli~\etal\pub{ECCV22}~\cite{eboli2022fast}   & \ding{51}      & Optimization\&CNN & Regression                                                                                                               \\
                      & PI$^2$RNet\pub{TCI23}~\cite{jiang2023annular}      & \ding{51}      & CNN              & Regression                                                                                                               \\
                      & Wei ~\etal\pub{OE24}~\cite{wei2024computational}    & \ding{51}      & CNN\&Transformer  & Regression                                                                                                               \\
                      & UniFMIR\pub{NM24}~\cite{ma2024pretraining}     & \ding{51}      & Transformer      & Regression                                                                                                               \\ \hline
\multirow{14}{*}{IR}  & DPIR\pub{TPAMI21}~\cite{zhang2021plug}         & \ding{55}      & Optimization\&CNN              & Regression                                                                                                               \\
                      & DWDN\pub{TPAMI22}~\cite{2022DWDN}          & \ding{55}      & CNN & Regression                                                                                                              \\ \arrayrulecolor{gray}\cdashline{2-5}\arrayrulecolor{black}
                      & SRN-Deblur\pub{CVPR18}~\cite{tao2018srndeblur}   & \ding{51}      & RNN          & Regression                                                                                                             \\
                      & DeblurGANv2\pub{CVPR18}~\cite{Kupyn_2019_ICCV}  & \ding{51}      & CNN              & GAN                                                                                                            \\
                      & MIMOUNet\pub{ICCV21}~\cite{cho2021rethinking}   & \ding{51}      & CNN              & Regression                                                                                                              \\
                      & SwinIR\pub{ICCVW21}~\cite{liang2021swinir}       & \ding{51}      & Transformer      & Regression                                                                                                               \\
                      & MPRNet\pub{CVPR21}~\cite{Zamir2021MPRNet}       & \ding{51}      & CNN              & Regression                                                                                                               \\
                      & NAFNet\pub{ECCV22}~\cite{chen2022simple}       & \ding{51}      & CNN              & Regression                                                                                                               \\
                      & Restomer\pub{CVPR22}~\cite{zamir2022restormer}     & \ding{51}      & Transformer      & Regression                                                                                                               \\
                      & Uformer\pub{CVPR22}~\cite{wang2022uformer}      & \ding{51}      & Transformer      & Regression                                                                                                               \\
                      & FeMaSR\pub{MM22}~\cite{chen2022real}      & \ding{51}      & Transformer            & GAN                                                                                                               \\ \arrayrulecolor{gray}\cdashline{2-5}\arrayrulecolor{black} 
                      & DRBNet\pub{CVPR22}~\cite{ruan2022learning}        & \ding{51}      & CNN              & Regression                                                                                                             \\ \arrayrulecolor{gray}\cdashline{2-5}\arrayrulecolor{black}
                      & PromptIR\pub{NeurIPS23}~\cite{potlapalli2023promptir}      & \ding{51}      & Transformer      & Regression                                                                                                            \\ \arrayrulecolor{gray}\cdashline{2-5}\arrayrulecolor{black} 
                      & DiffBIR\pub{ECCV24}~\cite{lin2024diffbir}      & \ding{51}      & CNN              & Diffusion                                                                                                                \\ \Xhline{2\arrayrulewidth} % 底部加粗横线（宽度为默认的2倍）
\end{tabular}%
}

%% file: Table/evaluate_all_revised.tex
\setlength\tabcolsep{3pt} % 设置表格列之间的间距
\small
\begin{tabular}{r||ccc||ccccccc}
% \hline
\Xhline{2\arrayrulewidth} % 底部加粗横线（宽度为默认的2倍）
\rowcolor{tableHeadGray}
Method       & Params (M) & FLOPs (G) & Time (ms)& PSNR$\uparrow$   & SSIM$\uparrow$   & LPIPS$\downarrow$   & FID$\downarrow$  & OIQE$\uparrow$   & ClipIQA$\uparrow$   & O. P.$\uparrow$ \\ \hline
Wiener\pub{MITPress49}~\cite{wiener1949extrapolation} & \raisebox{0.5ex}{\rule{0.3cm}{0.5pt}}  & \raisebox{0.5ex}{\rule{0.3cm}{0.5pt}} & 0.21 & 19.62     & 0.699     & 0.448      & 122.8    & 0.475  & 0.242        & 0.972 (24)      \\
FSANet\pub{OE22}~\cite{lin2022non}        & 12.86 & 54.14 & 7.57 & 21.60 & 0.751    & 0.268    & 88.19    & 0.298  & 0.344       & 1.191 (20)  \\
PART\pub{TIP24}~\cite{jiang2024minimalist}         & 19.35 & 51.86 & 79.09  & \textbf{28.10} & 0.866    & 0.228   & 43.66    & 0.608  & 0.389       & 1.494 (7)      \\ \arrayrulecolor{gray}\cdashline{1-11}\arrayrulecolor{black}
Peng ~\etal\pub{TOG19}~\cite{peng2019learned}  & 44.59 & 21.68 & 1.01   & 21.81    & 0.744     & 0.318      & 104.5    & 0.279 & 0.267    & 1.108 (21)      \\
DFUnet\pub{TOG21}~\cite{chen2021optical}       & 4.23   & 18.90 & 5.99 & 26.33    & 0.844     & 0.252      & 52.28    & 0.611 & 0.393    & 1.436 (12)      \\
FOV-KPN\pub{ICCV21}~\cite{chen2021extreme_quality}     & 4.01   & 19.00 & 9.68 & 26.34    & 0.824     & \underline{0.184}      & 50.27    & 0.631 & 0.422    & 1.502 (6)      \\
Eboli~\etal\pub{ECCV22}~\cite{eboli2022fast} & \raisebox{0.5ex}{\rule{0.3cm}{0.5pt}}  & \raisebox{0.5ex}{\rule{0.3cm}{0.5pt}} & 16.10 & 21.83    & 0.681     & 0.395      & 107.8    & 0.307      & 0.253     & 0.998 (23)      \\
PI$^2$RNet\pub{TCI23}~\cite{jiang2023annular}      & 14.47  & 128.33  & 80.77 & 25.28    & 0.851     & 0.251      & 59.17    & 0.651 & 0.361    & 1.434 (13)   \\
Wei ~\etal\pub{OE24}~\cite{wei2024computational}    & 9.83   & 48.81   & 7.72 & 25.51    & 0.841     & 0.273      & 62.81    & 0.546 & 0.350    & 1.371 (17)     \\
UniFMIR\pub{NM24}~\cite{ma2024pretraining}       & 1.65   & 99.71   & 109.78 & 24.01     & 0.799     & 0.313      & 89.83    & 0.406 & 0.342    & 1.225 (18)       \\ \hline
DPIR\pub{TPAMI21}~\cite{zhang2021plug}       & \raisebox{0.5ex}{\rule{0.3cm}{0.5pt}}     & \raisebox{0.5ex}{\rule{0.3cm}{0.5pt}} & 61.18 & 19.43    & 0.728  & 0.404      & 107.3    & 0.600      & 0.355   & 1.043 (22)      \\
DWDN\pub{TPAMI22}~\cite{2022DWDN}       & 7.05  & 86.13  & 28.44 & 25.00    & 0.824  & 0.270      & 61.85    & 0.634    & 0.348   & 1.388 (16)     \\ \arrayrulecolor{gray}\cdashline{1-10}\arrayrulecolor{black}
SRN-Deblur\pub{CVPR18}~\cite{tao2018srndeblur}  & 10.25 & 108.52 & 10.12 & 26.93   & 0.847  &  0.249     &  50.61   & 0.614     & 0.383   & 1.446 (11)      \\
DeblurGANv2\pub{CVPR18}~\cite{Kupyn_2019_ICCV} & 72.14 & 48.79  & 14.88 & 22.33   & 0.752  &  0.276     &  98.91   & 0.353     & 0.405   & 1.202 (19)     \\
MIMOUNet\pub{ICCV21}~\cite{cho2021rethinking}  & 16.11  & 153.93 & 12.14 & 27.36   & 0.870  &  0.229     &  45.05   & \textbf{0.742}     & 0.383   & 1.527 (4)      \\
SwinIR\pub{ICCVW21}~\cite{liang2021swinir}     & 11.97  & 50.84 & 59.91 & 27.22   & 0.848  &  0.264     &  54.41   & 0.513     & 0.370   & 1.398 (14)       \\
MPRNet\pub{CVPR21}~\cite{Zamir2021MPRNet}      & 20.13  & 1707.35 & 41.15 & 26.54   & 0.846  &  0.260     &  69.50   & 0.577     & 0.310   & 1.395 (15)      \\
NAFNet\pub{ECCV22}~\cite{chen2022simple}      & 67.89 & 63.05 & 0.64 & 27.78   & \textbf{0.876}   & 0.211      &  39.19   & 0.705     & 0.404   & \underline{1.549 (2)}      \\
Restomer\pub{CVPR22}~\cite{zamir2022restormer}  & 26.13  & 140.99 & 49.84 & 27.04    & 0.867  & 0.233      &  62.75   & 0.704     & 0.321   & 1.484 (8)      \\
Uformer\pub{CVPR22}~\cite{wang2022uformer}   & 50.47  & 85.78 & 13.07 & \underline{27.95}   & \underline{0.873}   & 0.220       &  52.81   & 0.714     & 0.369   & 1.525 (5)      \\
FeMaSR\pub{MM22}~\cite{chen2022real}      & 37.37  & 115.91 & 18.85 & 26.94   & 0.841   & \textbf{0.136}      & \textbf{34.59}     & \underline{0.722}     & \underline{0.520}   & \textbf{1.618 (1)}      \\ \arrayrulecolor{gray}\cdashline{1-11}\arrayrulecolor{black}
DRBNet\pub{CVPR22}~\cite{ruan2022learning}    & 11.69  & 49.16 & 4.68 & 26.82   & 0.851   & 0.254       &  60.69    & 0.673     & 0.344   & 1.447 (10)      \\ \arrayrulecolor{gray}\cdashline{1-11}\arrayrulecolor{black}
PromptIR\pub{NeurIPS23}~\cite{potlapalli2023promptir}  & 35.59 & 158.14  & 53.50& 26.96 & 0.862    & 0.241       & 64.74    & 0.719      & 0.312   &  1.474 (9)     \\ \arrayrulecolor{gray}\cdashline{1-11}\arrayrulecolor{black}
DiffBIR\pub{ECCV24}~\cite{lin2024diffbir}   & \raisebox{0.5ex}{\rule{0.3cm}{0.5pt}}      & 380.00 & 3168.01 & 27.65   & 0.812    & 0.196      & \underline{41.06}    & 0.711       & \textbf{0.623}  & 1.547 (3)      \\ \Xhline{2\arrayrulewidth} % 底部加粗横线（宽度为默认的2倍）
\end{tabular}%

%% file: sec_camera_ready/6_conclusion.tex
\section{Conclusion and Discussion}
To our knowledge, we establish the first universal CAC benchmark for photographic cameras, evaluating diverse image restoration and CAC models across dimensions.
A new benchmark named \ourdataset, containing a sufficient number of spherical and aspherical lenses that adhere to physical constraints, has been constructed by extending the recently advanced AOD method. 
To quantify CAC difficulty, we introduce ODE, a new aberration quantization framework that guides lens selection and improves data distribution.
Our comprehensive evaluation of 24 models leads to 9 key observations, from which we further propose several recommendations for advancing universal CAC:
\begin{itemize}
\item CNN-based architectures effectively correct most aberrations with reasonable inference time.
\item Diffusion-based models are well-suited for severe aberrations.
\item Incorporating optical and clean image priors significantly improves CAC performance.
\end{itemize}
This extensive benchmark, along with our analysis and findings, deepens the understanding of CAC and supports the development of more robust and generalizable methods.

%% file: sec_camera_ready/ack.tex
\section*{Acknowledgment}

This research was funded by the Natural Science Foundation of Zhejiang Province (Grant No. LZ24F050003), National Natural Science Foundation of China (Grant Nos. 12174341 and 62473139), the Ministry of Education and Science of Bulgaria, as part of the Bulgarian National Roadmap for Research Infrastructure (support for INSAIT) and the European Union's Horizon Europe -- the Framework Programme for Research and Innovation, under grant agreement 101168521. The research was also funded by Hunan Provincial Research and Development Project (Grant 2025QK3019) and State Key Laboratory of Autonomous Intelligent Unmanned Systems (the opening project number ZZKF2025-2-10).

%% file: supp.tex
In this supplementary material, 
\S\ref{sec:Comparison with Existing Benchmarks} describes a comparison between our benchmark and existing benchmarks.
\S\ref{sec:Implementation Details} provides additional implementation details on the datasets and benchmark, and also presents further simulation results compared with \textit{Zemax}, along with the generation process of the pseudo-GT.
\S\ref{sec:Optical Degradation Evaluator} describes the ODE calculation process and highlights its importance for CAC. \S\ref{sec:More CAC Results} provides detailed CAC results. \S\ref{sec:Validation of Our Observations} presents ablation studies to validate our key findings. 
\S\ref{sec:effectiveness of UNICAC} demonstrates the advantages of \ourdataset ~considering aspherical lens designs. 
Finally, \S~\ref{sec:Limitations and Future Work} discusses the limitations of our work and outlines directions for future research.

\begin{table*}[h]
    \centering
    \resizebox{\linewidth}{!}{%
    \begin{tabular}{l c c c l}
        \toprule
        \textbf{Work} & \textbf{Key Specs (FoV, F/\#, Elem.)} & \textbf{Test Set (\# Lenses)} & \textbf{Lens Type} & \textbf{Aberration Quantification} \\
        \midrule
        Li~\textit{et al.}~\cite{li2021universal} & Unspecified & Small / Unknown & Unspecified & N/A (Random) \\
        Gong~\textit{et al.}~\cite{gong2024physics} & Unspecified & $\sim$5 & Unspecified & N/A (Random) \\
        OmniLens~\cite{jiang2024flexible} & Discrete (6 Specific Sets) & 6 (Handcrafted) & Spherical + Aspherical & Spot RMS Radius \\
        \midrule
        \textbf{\ourdataset~(Ours)} & \textbf{40$\sim$80$^\circ$, F/2$\sim$4, 1$\sim$6 Elem.} & \textbf{120 (Unseen)} & \textbf{Spherical + Aspherical} & \textbf{ODE (Proposed)} \\
        \bottomrule
    \end{tabular}%
    }
    \vspace{2mm} % 
    \caption{Comparison with existing CAC benchmarks. \ourdataset~features strictly defined optical design ranges and a large-scale stratified test set, whereas prior works rely on unspecified or discrete lens samples.}
    \label{tab:supp_comparison}
\end{table*}

\begin{figure*}[!t]
  \centering
  \includegraphics[width=0.9\linewidth]{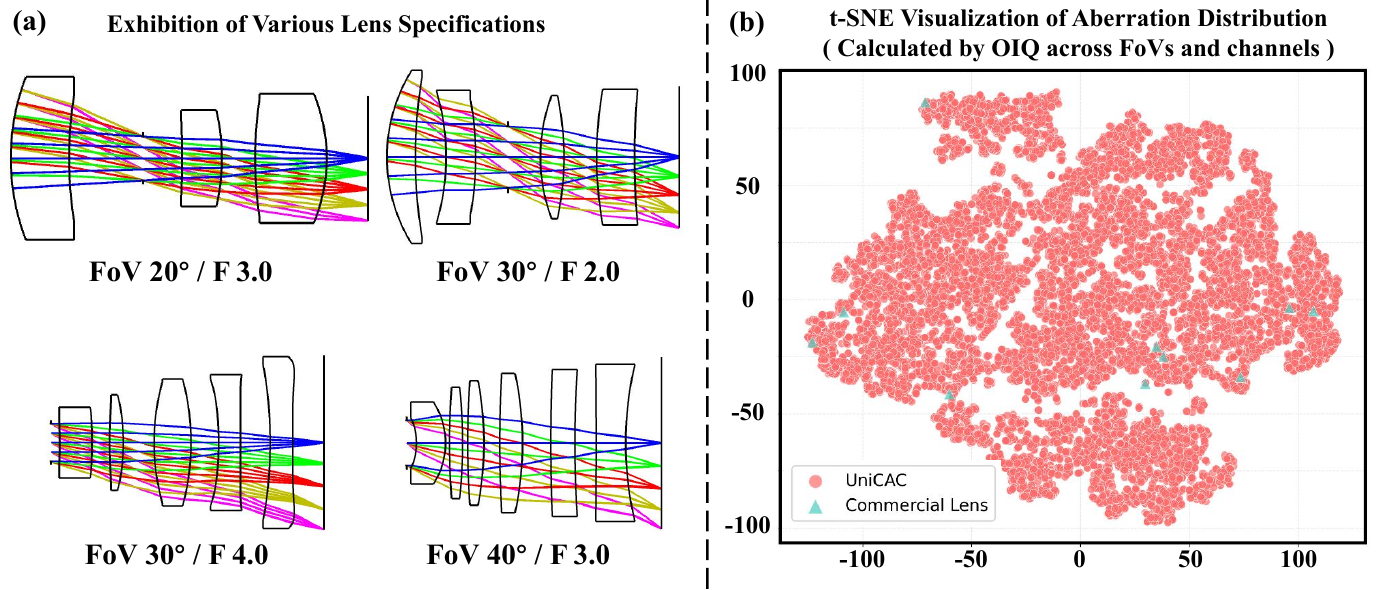}
  \caption{(a) Examples of automatically designed lenses with diverse specifications. (b) t-SNE visualization of aberration distributions in our benchmark. Best viewed when zoomed in.}
  \label{fig:re_benchmark_constrcution}
\end{figure*}

\section{Comparison with Existing Benchmarks}
\label{sec:Comparison with Existing Benchmarks}

To clearly highlight the scale, diversity, and rigorous physical definitions of our proposed dataset, we provide a detailed comparison with existing cross-lens CAC benchmarks in Table~\ref{tab:supp_comparison}. Compared to prior works that rely on small-scale, handcrafted, or unspecified lens samples, \ourdataset~features a large-scale stratified test set with strictly defined optical design ranges and introduces the ODE for objective aberration quantification.

\section{More Implementation Details}
\label{sec:Implementation Details}
\subsection{Dataset Details}
The key to constructing a dataset is to ensure the diversity of lens aberrations. 
To generate a credible benchmark, we sampled based on the proposed ODE, ensuring a reasonable distribution of CAC difficulty.
In terms of training and validation data, we used $873$ and $75$ lens images, respectively, to simulate aberration images on the DIV2K and Flickr2K~\cite{timofte2017ntire} datasets.
For the test set, we select additional $120$ lenses, ensuring that the imaging quality, uniformity, and chromatic aberration distribution are reasonable.
The GT images of the test set are taken with a Sony $\alpha$6600 camera and a Sony $18{-}135mm$ lens, with a total of $26$ images. 
We apply the test lens to the $26$ images one by one for aberration simulation.
Specifically, regarding the image formation model, we simulate the broadband degradation of the optics by densely ray-tracing across the continuous visible spectrum (\textit{e.g.}, $400\text{nm}$ to $700\text{nm}$). These wavelength-dependent PSFs are subsequently integrated with a standard camera response function to convert the continuous spectral degradations into highly accurate RGB kernels, ensuring maximum physical fidelity to real-world sensors. 
Furthermore, our simulation explicitly integrates the full ISP pipeline, including Bayer patterning and demosaicking. To bridge the sim-to-real gap, we apply synchronized perturbations (primarily white balance gains) to both LQ and GT images as ISP domain randomization. This strategy forces the model to learn aberration features that are invariant to specific color renderings. 
According to the research needs, these $120$ lenses are reasonably selected for analysis to ensure the comprehensiveness and accuracy of the research results.

\begin{figure*}[!t]
  \centering
  \includegraphics[width=0.9\linewidth]{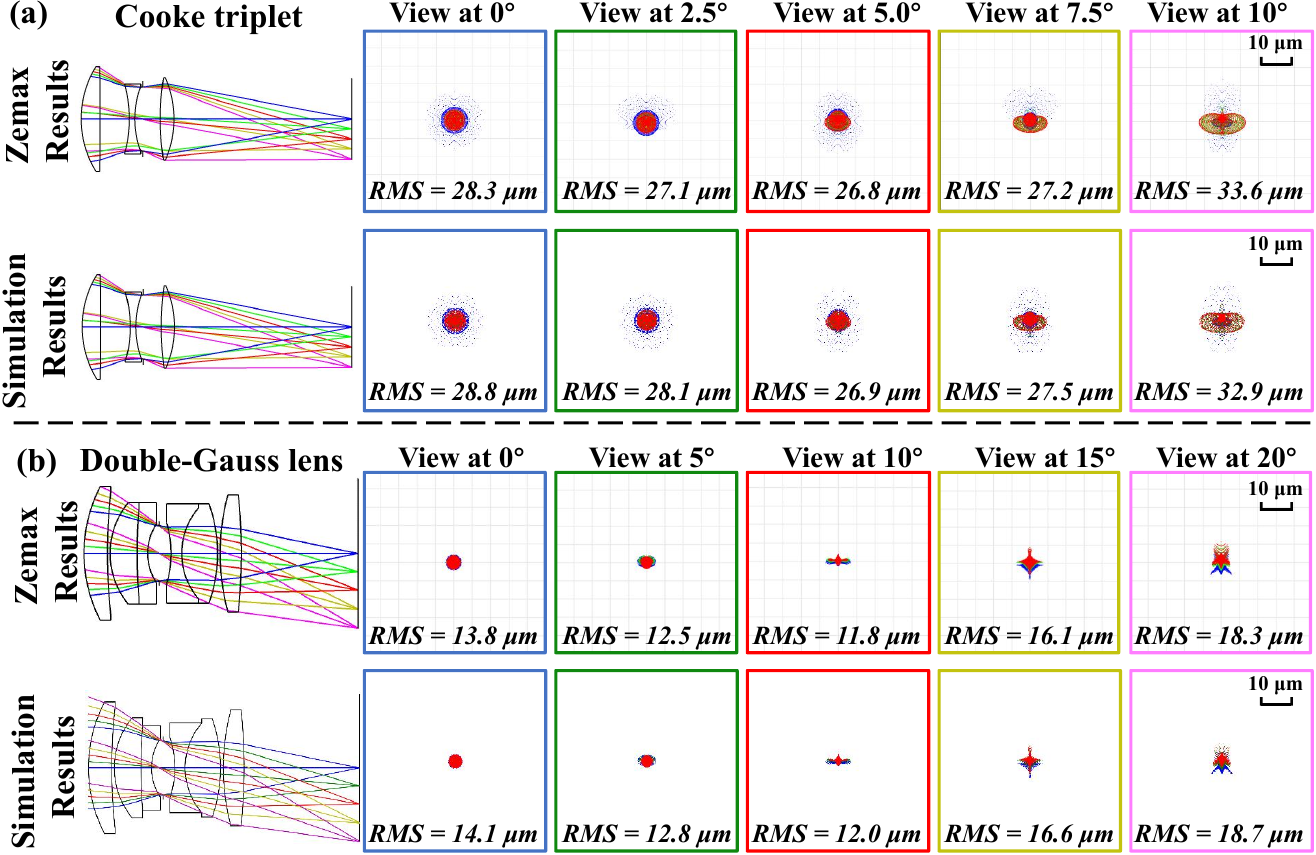}
  \caption{More comparison with \textit{Zemax}. For additional validation, we test the simulation system using Cooke triplet lens and Double-Gauss lens.}
  \label{fig:compare zmx rms}
\end{figure*}

\subsection{Lens Generation Details}
OptiFusion~\cite{gao2024global} is an automatic design algorithm that generates physically-constrained optical lenses based on specified design requirements. 
It integrates Genetic Algorithm (GA), Simulated Annealing Algorithm (SAA), and ADAM optimization.
We first construct a spherical lens dataset based on OptiFusion.
The diversity of design specifications ensures that a variety of aberrations are present in our benchmark.
The value ranges for design specifications are: $1{\sim}6$ for piece number with an interval of $1$, $20^\circ{\sim}40^\circ$ for half FoV with an interval of $10^\circ$, and $2.0{\sim}4.0$ for F number with an interval of $1.0$. 
Fig.~\ref{fig:re_benchmark_constrcution}(a) presents several representative lenses with different specifications included in our benchmark.
Notably, the aperture position is determined based on the piece number, which can be located before the first piece, after the last piece, or in the middle of the lens. With all sets of specifications fed into OptiFusion, the initial spherical lens dataset is constructed.

Moreover, we extend OptiFusion's spherical surface parameters to encompass aspherical surfaces.
Given a Cartesian coordinate system $(x,y,z)$, the $z$-axis coincides with the optical axis, while $(x,y)$ forms the transverse plane. 
Let $r=\sqrt{x^2+y^2}$ and $\rho=r^2$, then the height $h$ of the traditional spherical surface is defined as:

\begin{equation}
\label{eq:gaoyao4}
    h(\rho) = \frac{c\rho}{1+\sqrt{1-c^2\rho}},
\end{equation}
where $c$ is the curvature. Further, the height of the standard aspherical surface is defined as:

\begin{equation}
\label{eq:gaoyao5}
    h(\rho) = \frac{c\rho}{1+\sqrt{1-\alpha\rho}}+\sum^n_{i=2}a_{2i}\rho^i,
\end{equation}
where $\alpha=(1+k)c^2$ with $k$ being the conic coefficient, and $a_{2i}$'s are higher-order coefficients. 
By setting the conic coefficient and aspherical coefficients as variables, we continue to use OptiFusion to further optimize all the lenses in the initial spherical lens dataset, so as to establish an additional aspherical lens dataset.
Combining the initial spherical lens dataset and the aspherical lens dataset, we can achieve a large lens library as our lens data foundation. 
It is worth noting that the automatically designed lenses conform to physical constraints.

\begin{figure}[!t]
  \centering
  \includegraphics[width=1.0\linewidth]{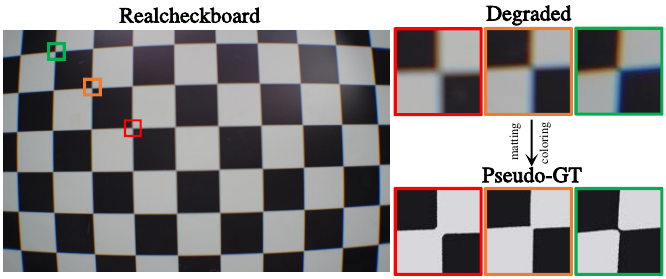}
  \caption{Comparison between the real aberration image and the extracted pseudo-GT image.} 
  \label{fig:checker}
\end{figure}

\begin{figure*}[!t]
  \centering
  \includegraphics[width=0.9\linewidth]{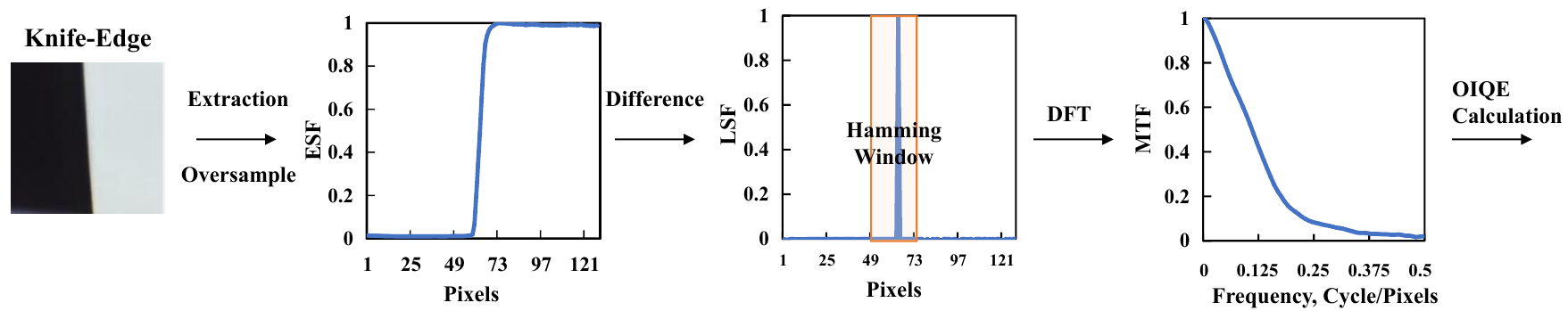}
  \caption{Pipeline for OIQE calculation.} 
  \label{fig:oiqe}
\end{figure*}
\subsection{Lens Aberration Distribution Analysis}
To validate the effectiveness and reliability of our benchmark, we perform a t-SNE analysis on the aberration distribution of the benchmark lenses. Following the approach in~\cite{gong2024physics}, we select several well-known commercial lenses from ZeBase as references. 
As shown in Fig.~\ref{fig:re_benchmark_constrcution}(b), the aberration distribution in our benchmark is broad and diverse, effectively covering the range of common aberrations found in real-world optical systems.

\subsection{Aberration Simulation Evaluation Details}
\label{sec:Aberration Simulation Evaluation}
\noindent\textbf{More comparison with \textit{Zemax}}:
In Fig.~\ref{fig:compare zmx rms}, we present additional comparison results for multi-element optical systems to further validate the fidelity of our simulation system.
The minor discrepancies mainly stem from three aspects: ray tracing accuracy, the precision of entrance pupil positioning, and the method of ray sampling at the entrance pupil.

\noindent\textbf{Pseudo-GT}:
Obtaining pixel-aligned Ground Truth (GT) in the validation of aberration simulation and real-world photography is challenging. According to~\cite{chen2021extreme_quality}, we generate pseudo-GT by performing matting and coloring on aberration images, shown in Fig.~\ref{fig:checker}. However, the edges in the pseudo-GT differ from those in the real GT due to severe aberrations causing errors in the edge detection operators.
Given this situation, we should focus more on the visual effects of aberrations caused by the PSF as demonstrated in Fig.~4 in the main text.

\begin{figure*}[t]
  \centering
  \includegraphics[width=0.95\linewidth]{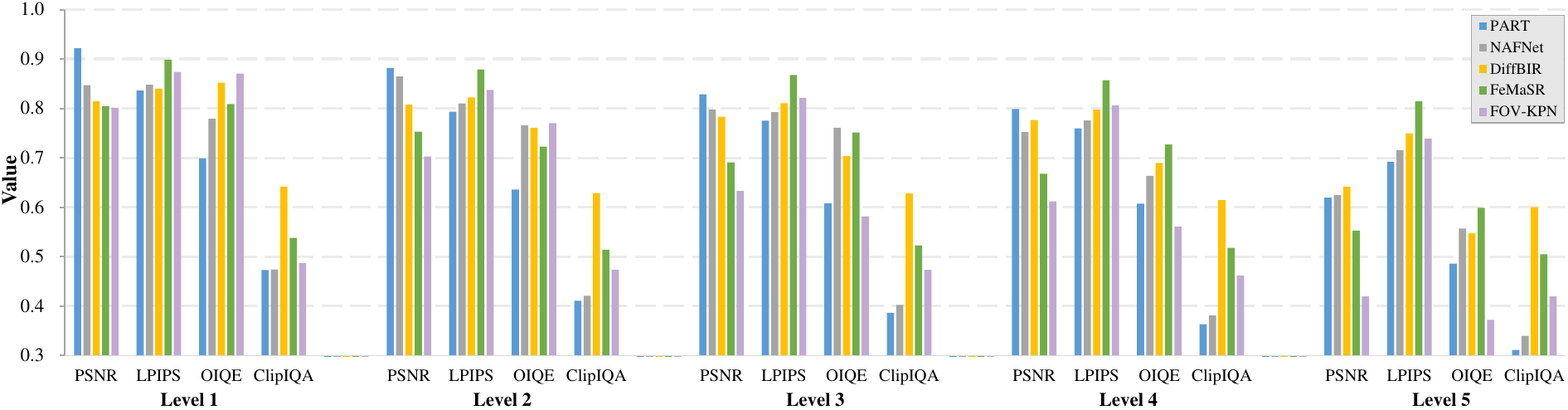}
  \caption{\textbf{Overall performance comparison of different methods across five aberration severity levels.} To enhance visual clarity, we normalize PSNR to a range of $20dB$ to $30dB$ and transform LPIPS values by subtracting them from 1.} 
  \label{fig:level compare zzt}
\end{figure*}

\begin{figure*}[!t]
  \centering
  \includegraphics[width=1.0\linewidth]{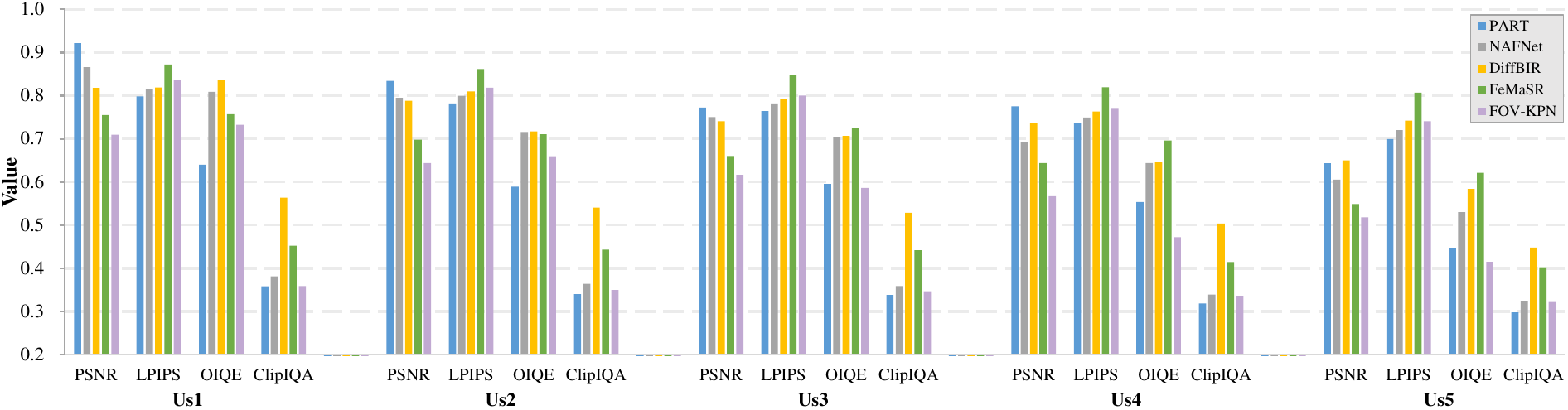}
  \caption{Overall performance comparison of different methods under $5$ image spatial uniformity levels.} 
  \label{fig:uniformity compare zzt}
\end{figure*}

\begin{figure*}[h]
  \centering
  \includegraphics[width=0.95\linewidth]{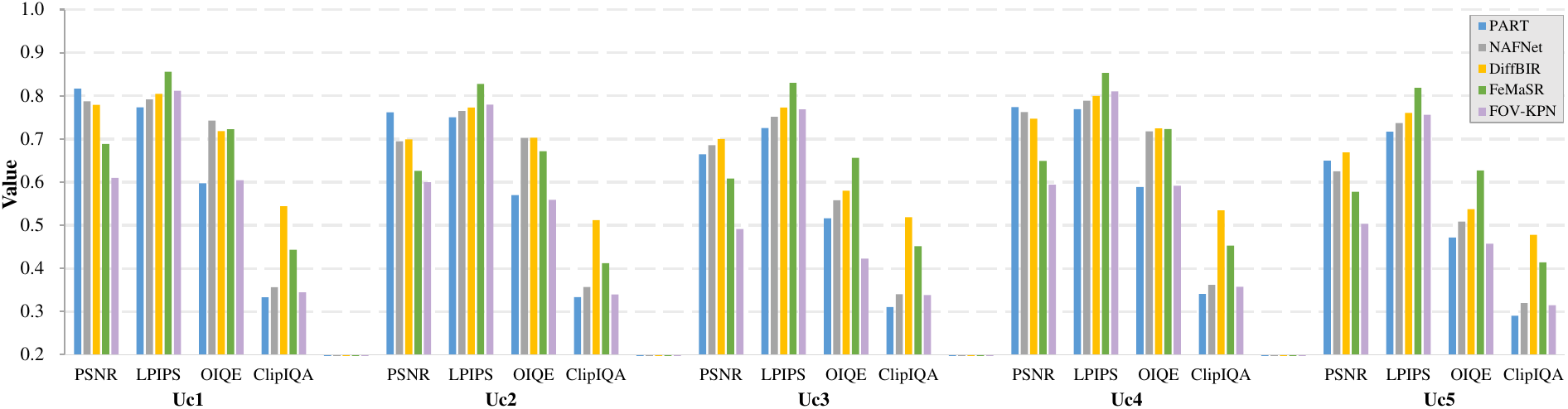}
  \caption{Overall performance comparison of different methods under $5$ image chromatic aberration levels.} 
  \label{fig:chromatic aberration compare zzt}
\end{figure*}

\section{More Discussion on the Optical Degradation Evaluator}
\label{sec:Optical Degradation Evaluator}

\subsection{Calculation Details}
As introduced in the main paper, we compute the Optical Degradation Evaluator (ODE) by first calculating sub-OIQ values across five fields of view (FoVs) and three color channels for a degraded checkerboard. The overall OIQ is then obtained by averaging these values.

For $U_s$, we merge channel data per FoV and compute the coefficient of variation (CV) across FoVs.
For $U_c$, we combine FoV data for each channel and calculate the CV across channels.
Additionally, we present the specific computation details of OIQE~\cite{jiang2024minimalist}, which is a key component of ODE in Fig.~\ref{fig:oiqe}.
After calculating the MTF using the edge method, OIQE is obtained by averaging MTF50 and MTF Area.
The CV is calculated as follows:
\begin{equation}
    CV= \frac{Std(\left \{ OIQ_{i} | i=1,...,N \right \}  )}{Avg(\left \{ OIQ_{i} | i=1,...,N \right \})}.
    \label{eq:CV}
\end{equation}
\subsection{Importance of ODE}
Traditional metrics such as MTF or Strehl ratio focus solely on optical performance, whereas \textbf{ODE} combines both optical quality (through MTF-based OIQE) and image fidelity (via PSNR and SSIM), providing a more comprehensive quantification of CAC difficulty. 
\textbf{ODE} demonstrates \textit{a strong linear relationship with restoration quality}, offering valuable guidance for computational imaging researchers in selecting appropriate lenses.

Beyond replacing the \textbf{RMS radius}, \textbf{ODE} also \textit{bridges the gap between optical design and CAC}, offering unique insights to guide optical engineers. 
In practical scenarios, optical design is often constrained by factors like cost, volume, weight, or manufacturing feasibility, making it difficult to design optical systems with optimal performance. 
\textbf{ODE}, \textit{a framework for quantifying CAC difficulty}, empowers designers to make targeted trade-offs.

\section{More CAC Results}
\label{sec:More CAC Results}
\noindent\textbf{Quantitative results under $5$ different aberration severity levels.}
We systematically demonstrate the quantitative metrics of $10$ methods across various levels of aberration severity in Tab.~\ref{tab:supp level compare}.
All methods show a decline in overall performance as the severity of aberration increases, indicating a strong correlation between CAC results and ODE.
The results indicate that the PART~\cite{jiang2024minimalist}, which leverages PSF information, achieves the highest PSNR from level L1 to L4. 
However, in the most severe aberration level (L5), the diffusion-based DiffBIR~\cite{lin2024diffbir} outperforms others, achieving the best results. 
When focusing on perceptual metrics, methods like FeMaSR~\cite{chen2022real} and DiffBIR~\cite{lin2024diffbir}, which utilize GAN or diffusion training paradigms, exhibit significant advantages over other approaches.
Additionally, we select five representative networks to provide a more detailed comparison of the specific metric differences across different severity levels, as illustrated in Fig~\ref{fig:level compare zzt}.

\noindent\textbf{Quantitative results under $5$ different spatial uniformity levels.}
The detailed metrics are presented in Tab.~\ref{tab:supp uf compare} under different levels of spatial uniformity.
Spatial uniformity also affects CAC results, as network performance declines with increasing spatial non-uniformity.
To better illustrate metric variation across spatial uniformity levels, Fig.~\ref{fig:uniformity compare zzt} presents results from five representative methods.

\noindent\textbf{Quantitative results under $5$ different chromatic aberration levels.}
Given that chromatic aberration has a minimal impact on CAC results, we have showcased the performance of five representative methods across different chromatic aberration levels. 
As illustrated in Fig.~\ref{fig:chromatic aberration compare zzt}, the model's CAC performance exhibits a low correlation with levels of chromatic aberration, indicating that it is not a primary factor influencing CAC results.

\section{Validation of Our Observations.}
\label{sec:Validation of Our Observations}
To verify whether our key observations can provide actionable guidance for CAC model design, we conduct a series of controlled experiments by making targeted modifications inspired by the insights.
This not only helps validate the observations themselves but also demonstrates their practical value in improving CAC performance.
In Tab.~\ref{tab:abaltion_fema}, we evaluate the effectiveness of the training strategy and the incorporation of clear image priors. The GAN-based training method improves perceptual quality, while the clear image prior enhances both image fidelity and perceptual realism.

As illustrated in Tab.~\ref{tab:ablation_optics}, incorporating FoV and PSF information effectively improves the model’s CAC performance in terms of both image fidelity and perceptual quality.
\begin{table}[t]

    \centering  
    \captionsetup{font={normal}} 
    \setlength{\tabcolsep}{8pt}
    \renewcommand\arraystretch{1.1}
    \resizebox{0.48\textwidth}{!}{%
        \begin{tabular}{c||cc||ccc}
        \hline

        Method                  & GAN & Codebook & PSNR$\uparrow$  & LPIPS$\downarrow$ & ClipIQA$\uparrow$ \\ \hline
        \multirow{3}{*}{FeMaSR} & \ding{55}   & \ding{55}        & 26.83 & 0.147 & 0.491   \\
                                & \ding{51}   & \ding{55}        & 26.86 & 0.145 & 0.512   \\
                                & \ding{51}   & \ding{51}        & \textbf{26.94} & \textbf{0.136} & \textbf{0.520}   \\ \hline
        \end{tabular}}

\caption{Ablation results of training scheme and clean image prior}
\label{tab:abaltion_fema}  
\end{table}

\begin{table}[t]  

    \centering 
    \captionsetup{font={normal}} 
    \setlength{\tabcolsep}{12pt}
    \renewcommand\arraystretch{1.1}
    \resizebox{0.48\textwidth}{!}{%

        \begin{tabular}{c||c||ccc}
        \hline
        Method                  & FoV & PSNR$\uparrow$  & LPIPS$\downarrow$ & ClipIQA$\uparrow$ \\ \hline
        \multirow{2}{*}{DFUnet} & \ding{55}   & 25.88 & 0.253 & 0.377   \\
                                & \ding{51}   & \textbf{26.33} & \textbf{0.252} & \textbf{0.393}   \\ \hline
        Method                  & PSF & PSNR$\uparrow$  & LPIPS$\downarrow$ & ClipIQA$\uparrow$ \\ \hline
        \multirow{2}{*}{PART}   & \ding{55}   & 27.22 & 0.264 & 0.370   \\
                                & \ding{51}   & \textbf{28.10} & \textbf{0.228} & \textbf{0.389}    \\ \hline
        \end{tabular}}

\caption{Ablation results of optical prior}  
\label{tab:ablation_optics}  
\end{table}

\begin{table*}[!t]
\captionsetup{font=normal}
\centering
\input{Table/lib_compare}

\caption{Quantitative evaluation of the \ourdataset.}
\label{tab:lib_compare}
\end{table*}

\section{On the Effectiveness of \ourdataset}
\label{sec:effectiveness of UNICAC}
To demonstrate the effectiveness of our dataset in considering both spherical and aspherical lenses, we compare it with another dataset based on automatic design that only considers spherical lenses~\cite{jiang2024flexible}.
All experiments are conducted under the same experimental environment and settings.
As proven in Tab.~\ref{tab:lib_compare}, our \ourdataset~outperforms AODLib because it incorporates aspherical lenses into the overall pipeline.

\section{Limitations and Future Work}
\label{sec:Limitations and Future Work}
While our comprehensive benchmark yields valuable insights, several areas remain open for further improvement.
First, although the extended AOD algorithm~\cite{gao2024global} has automatically designed a sufficient number of spherical and aspherical lenses, our benchmark primarily focuses on photographic cameras. We have not yet considered advanced optical components such as wavefront-based metasurfaces and Diffractive Optical Elements (DOEs), nor does it include other imaging devices like microscopes, telescopes, and endoscopes, which also suffer from aberrations but with an emphasis on different types. Future research can further incorporate these complex optical systems into the lens library to enhance the overall dataset's aberration diversity.
Furthermore, the aberration images used in this study are based on optical simulation~\cite{chen2021optical}. To isolate intrinsic lens aberrations, we currently fix the object distance at infinity, reserving depth-dependent PSFs for future work. Although the simulation accuracy has been validated, future work can incorporate depth-coupled degradations and include real-world images captured using various types of lenses to enhance the comprehensiveness and effectiveness of the dataset.

In the future, we will extend existing AOD algorithms to automatically design a wider variety of optical systems, enhancing the diversity of our dataset. Additionally, it will be essential to develop a more general and powerful CAC architecture based on this enriched dataset.

\begin{table*}[!t]
\captionsetup{font=normal}
\centering
\input{Table/supp_10method_5level_compare_revised}
\caption{Quantitative results under $5$ different aberration severity levels.}
\label{tab:supp level compare}
\end{table*}

\begin{table*}[!t]
\captionsetup{font=normal}
\centering
\input{Table/supp_10method_5Uf_compare_revised}
\caption{Quantitative results under $5$ different spatial uniformity levels.}
\label{tab:supp uf compare}
\end{table*}

%% file: Table/lib_compare.tex
\small
\setlength\tabcolsep{5pt} % 设置表格列之间的间距
\renewcommand\arraystretch{1.0} % 设置表格行之间的间距  
\setlength{\tabcolsep}{12pt}
\renewcommand\arraystretch{1.1}
\resizebox{0.9\textwidth}{!}{%
\begin{tabular}{cr||cccccc}
% \bottomrule[0.15em]
\hline
% \rowcolor{tableHeadGray}
             CAC Dataset                    &          Model              & PSNR$\uparrow$                      & SSIM$\uparrow$                      & LPIPS$\downarrow$                     & FID$\downarrow$                        & OIQE$\uparrow$ & ClipIQA$\uparrow$ \\ \hline
\multirow{3}{*}{AODLib~\cite{jiang2024flexible}}          & FOV-KPN\pub{ICCV21}~\cite{chen2021extreme_quality}  & 23.90 & 0.756 & 0.244 & 74.51  & 0.477 & 0.430   \\
                                 & SwinIR\pub{ICCVW21}~\cite{liang2021swinir}                 & 25.05 & 0.793 & \textbf{0.226} & 65.18  & 0.418 & \textbf{0.418}   \\
                                 & FeMaSR\pub{MM22}~\cite{chen2022real}                 & 24.15 & 0.748 & 0.235 & 67.73  & 0.568 & 0.476   \\ \arrayrulecolor{gray}\cdashline{1-8}\arrayrulecolor{black}
\multirow{3}{*}{UniCAC (Ours)} & FOV-KPN\pub{ICCV21}~\cite{chen2021extreme_quality}                & \textbf{26.34} & \textbf{0.824} & \textbf{0.184} & \textbf{50.27}  & \textbf{0.631} & \textbf{0.463}   \\
                                 & SwinIR\pub{ICCVW21}~\cite{liang2021swinir}                 & \textbf{27.22} & \textbf{0.848} & 0.264 & \textbf{54.41}  & \textbf{0.513} & 0.370   \\
                                 & FeMaSR\pub{MM22}~\cite{chen2022real}                 & \textbf{26.94} & \textbf{0.841} & \textbf{0.136} & \textbf{34.59}  & \textbf{0.722} & \textbf{0.520}   \\ \hline
\end{tabular}}

%% file: Table/supp_10method_5level_compare_revised.tex
\setlength\tabcolsep{4pt} % 减小列间距
\small % 缩小字体增强紧凑性
\begin{tabular}{l|c|c|c|c|c|c|c|c|c|c|c} % 12列格式
\hline
\multicolumn{1}{c|}{\multirow{2}{*}{Metric}} & 
\multicolumn{1}{c|}{\multirow{2}{*}{\begin{tabular}[c]{@{}c@{}}Aberration \\ Severity\end{tabular}}} & 
\multicolumn{10}{c}{Model} \\ \cline{3-12} 
\multicolumn{1}{c|}{} & \multicolumn{1}{c|}{} & PART & DFUnet & FOV-KPN & MIMOUnet & NAFNet & Restomer & Uformer & FeMaSR & PromptIR & DiffBIR \\ \hline
\multirow{5}{*}{PSNR$\uparrow$} & L1 & 29.22 & 27.57 & 28.02 & 28.24 & 28.47 & 27.77 & 28.87 & 28.05 & 27.77 & 28.14 \\ 
 & L2 & 28.82 & 27.07 & 27.03 & 28.22 & 28.65 & 27.77 & 28.66 & 27.53 & 27.63 & 28.08 \\ 
 & L3 & 28.29 & 26.45 & 26.33 & 27.52 & 27.98 & 27.33 & 28.10 & 26.91 & 27.35 & 27.84 \\ 
 & L4 & 27.99 & 25.81 & 26.12 & 27.39 & 27.53 & 27.04 & 27.98 & 26.68 & 26.94 & 27.77 \\ 
 & L5 & 26.19 & 24.75 & 24.20 & 25.43 & 26.25 & 25.29 & 26.12 & 25.53 & 25.12 & 26.42 \\ \hline
\multirow{5}{*}{SSIM$\uparrow$} & L1 & 0.907 & 0.895 & 0.881 & 0.912 & 0.915 & 0.906 & 0.911 & 0.885 & 0.903 & 0.830 \\ 
 & L2 & 0.881 & 0.860 & 0.842 & 0.885 & 0.889 & 0.882 & 0.886 & 0.855 & 0.878 & 0.821 \\ 
 & L3 & 0.868 & 0.843 & 0.823 & 0.871 & 0.877 & 0.871 & 0.875 & 0.841 & 0.866 & 0.815 \\ 
 & L4 & 0.858 & 0.831 & 0.810 & 0.860 & 0.867 & 0.858 & 0.865 & 0.830 & 0.853 & 0.809 \\ 
 & L5 & 0.818 & 0.792 & 0.765 & 0.819 & 0.830 & 0.817 & 0.826 & 0.796 & 0.811 & 0.784 \\ \hline
\multirow{5}{*}{LPIPS$\downarrow$} & L1 & 0.164 & 0.171 & 0.126 & 0.159 & 0.152 & 0.176 & 0.162 & 0.101 & 0.180 & 0.160 \\ 
 & L2 & 0.206 & 0.226 & 0.162 & 0.204 & 0.190 & 0.209 & 0.199 & 0.121 & 0.216 & 0.177 \\ 
 & L3 & 0.224 & 0.252 & 0.178 & 0.226 & 0.207 & 0.226 & 0.216 & 0.132 & 0.234 & 0.189 \\ 
 & L4 & 0.240 & 0.269 & 0.194 & 0.243 & 0.224 & 0.246 & 0.232 & 0.143 & 0.255 & 0.202 \\ 
 & L5 & 0.307 & 0.341 & 0.261 & 0.314 & 0.284 & 0.309 & 0.293 & 0.186 & 0.322 & 0.250 \\ \hline
\multirow{5}{*}{FID$\downarrow$} & L1 & 26.245 & 31.457 & 29.238 & 27.651 & 23.464 & 48.498 & 36.715 & 23.092 & 50.689 & 32.741 \\ 
 & L2 & 33.112 & 40.898 & 39.963 & 34.338 & 30.221 & 52.635 & 43.093 & 28.540 & 53.165 & 34.505 \\ 
 & L3 & 38.016 & 46.582 & 43.220 & 38.394 & 34.174 & 54.949 & 48.039 & 31.982 & 56.094 & 38.269 \\ 
 & L4 & 48.594 & 56.814 & 56.357 & 48.930 & 43.376 & 66.733 & 57.585 & 37.809 & 67.738 & 42.851 \\ 
 & L5 & 72.351 & 85.647 & 82.588 & 75.949 & 64.716 & 90.950 & 78.637 & 51.548 & 96.022 & 56.908 \\ \hline
\multirow{5}{*}{OIQE$\uparrow$} & L1 & 0.700 & 0.766 & 0.871 & 0.859 & 0.779 & 0.806 & 0.826 & 0.809 & 0.825 & 0.853 \\ 
 & L2 & 0.636 & 0.642 & 0.770 & 0.820 & 0.766 & 0.774 & 0.755 & 0.723 & 0.782 & 0.761 \\ 
 & L3 & 0.608 & 0.620 & 0.582 & 0.776 & 0.761 & 0.731 & 0.746 & 0.751 & 0.755 & 0.704 \\ 
 & L4 & 0.607 & 0.596 & 0.561 & 0.715 & 0.664 & 0.684 & 0.696 & 0.727 & 0.702 & 0.690 \\ 
 & L5 & 0.486 & 0.433 & 0.372 & 0.540 & 0.557 & 0.527 & 0.546 & 0.599 & 0.532 & 0.548 \\ \hline
\multirow{5}{*}{ClipIQA$\uparrow$} & L1 & 0.473 & 0.469 & 0.487 & 0.477 & 0.475 & 0.392 & 0.438 & 0.538 & 0.387 & 0.642 \\ 
 & L2 & 0.411 & 0.412 & 0.474 & 0.406 & 0.421 & 0.341 & 0.388 & 0.496 & 0.331 & 0.629 \\ 
 & L3 & 0.387 & 0.391 & 0.474 & 0.381 & 0.403 & 0.321 & 0.369 & 0.489 & 0.312 & 0.629 \\ 
 & L4 & 0.364 & 0.371 & 0.462 & 0.352 & 0.381 & 0.297 & 0.346 & 0.480 & 0.283 & 0.615 \\ 
 & L5 & 0.311 & 0.325 & 0.420 & 0.302 & 0.340 & 0.255 & 0.301 & 0.416 & 0.244 & 0.600 \\ \hline
\end{tabular}

%% file: Table/supp_10method_5Uf_compare_revised.tex
\setlength\tabcolsep{4pt} % 减小列间距
\small % 缩小字体增强紧凑性
\begin{tabular}{l|c|c|c|c|c|c|c|c|c|c|c} % 12列格式
\hline
\multicolumn{1}{c|}{\multirow{2}{*}{Metric}} & 
\multicolumn{1}{c|}{\multirow{2}{*}{\begin{tabular}[c]{@{}c@{}}Spatial \\ Uniformity\end{tabular}}} & 
\multicolumn{10}{c}{Model} \\ \cline{3-12} 
\multicolumn{1}{c|}{} & \multicolumn{1}{c|}{} & PART & DFUnet & FOV-KPN & MIMOUnet & NAFNet & Restomer & Uformer & FeMaSR & PromptIR & DiffBIR \\ \hline
\multirow{5}{*}{PSNR$\uparrow$} & Us1 & 29.22 & 26.87 & 27.10 & 28.30 & 28.66 & 27.98 & 28.89 & 27.56 & 27.78 & 28.18 \\ 
 & Us2 & 27.89 & 26.10 & 26.44 & 27.56 & 27.95 & 27.22 & 28.17 & 26.98 & 27.07 & 27.89 \\ 
 & Us3 & 27.73 & 25.97 & 26.17 & 27.07 & 27.50 & 26.76 & 27.74 & 26.61 & 26.69 & 27.41 \\ 
 & Us4 & 27.37 & 25.61 & 25.67 & 26.71 & 26.92 & 26.62 & 27.42 & 26.44 & 26.64 & 27.37 \\ 
 & Us5 & 26.43 & 24.75 & 25.19 & 25.65 & 26.06 & 25.42 & 26.11 & 25.49 & 25.35 & 26.51 \\ \hline
\multirow{5}{*}{SSIM$\uparrow$} & Us1 & 0.885 & 0.865 & 0.846 & 0.890 & 0.894 & 0.887 & 0.891 & 0.859 & 0.882 & 0.830 \\ 
 & Us2 & 0.826 & 0.849 & 0.834 & 0.877 & 0.883 & 0.874 & 0.879 & 0.851 & 0.870 & 0.826 \\ 
 & Us3 & 0.865 & 0.843 & 0.824 & 0.868 & 0.873 & 0.865 & 0.871 & 0.843 & 0.861 & 0.818 \\ 
 & Us4 & 0.811 & 0.827 & 0.811 & 0.851 & 0.858 & 0.850 & 0.855 & 0.831 & 0.846 & 0.811 \\ 
 & Us5 & 0.829 & 0.806 & 0.790 & 0.827 & 0.836 & 0.828 & 0.831 & 0.814 & 0.822 & 0.798 \\ \hline
\multirow{5}{*}{LPIPS$\downarrow$} & Us1 & 0.202 & 0.220 & 0.163 & 0.199 & 0.184 & 0.206 & 0.195 & 0.128 & 0.215 & 0.181 \\ 
 & Us2 & 0.190 & 0.240 & 0.181 & 0.219 & 0.200 & 0.225 & 0.212 & 0.138 & 0.232 & 0.190 \\ 
 & Us3 & 0.235 & 0.262 & 0.199 & 0.238 & 0.218 & 0.243 & 0.228 & 0.152 & 0.250 & 0.207 \\ 
 & Us4 & 0.236 & 0.285 & 0.228 & 0.270 & 0.251 & 0.273 & 0.259 & 0.180 & 0.280 & 0.236 \\ 
 & Us5 & 0.300 & 0.320 & 0.259 & 0.311 & 0.279 & 0.303 & 0.293 & 0.193 & 0.314 & 0.258 \\ \hline
\multirow{5}{*}{FID$\downarrow$} & Us1 & 36.281 & 44.166 & 41.581 & 35.766 & 32.920 & 54.721 & 45.589 & 30.527 & 55.965 & 36.512 \\ 
 & Us2 & 38.638 & 48.958 & 47.971 & 42.362 & 37.217 & 60.577 & 51.895 & 33.350 & 62.155 & 38.638 \\ 
 & Us3 & 48.667 & 58.617 & 57.240 & 51.937 & 43.631 & 66.133 & 57.538 & 38.260 & 69.144 & 43.284 \\ 
 & Us4 & 54.127 & 69.734 & 70.895 & 66.125 & 60.343 & 82.859 & 73.699 & 50.195 & 83.771 & 54.127 \\ 
 & Us5 & 65.855 & 77.619 & 69.387 & 73.627 & 61.036 & 84.739 & 76.640 & 48.737 & 87.640 & 56.686 \\ \hline
\multirow{5}{*}{OIQE$\uparrow$} & Us1 & 0.640 & 0.665 & 0.733 & 0.823 & 0.809 & 0.775 & 0.743 & 0.757 & 0.745 & 0.836 \\ 
 & Us2 & 0.718 & 0.555 & 0.660 & 0.766 & 0.716 & 0.696 & 0.674 & 0.711 & 0.722 & 0.718 \\ 
 & Us3 & 0.596 & 0.599 & 0.587 & 0.701 & 0.705 & 0.654 & 0.671 & 0.726 & 0.695 & 0.707 \\ 
 & Us4 & 0.646 & 0.528 & 0.472 & 0.644 & 0.644 & 0.583 & 0.623 & 0.696 & 0.625 & 0.646 \\ 
 & Us5 & 0.447 & 0.429 & 0.415 & 0.540 & 0.530 & 0.540 & 0.484 & 0.621 & 0.560 & 0.584 \\ \hline
\multirow{5}{*}{ClipIQA$\uparrow$} & Us1 & 0.359 & 0.367 & 0.359 & 0.364 & 0.382 & 0.325 & 0.350 & 0.452 & 0.315 & 0.564 \\ 
 & Us2 & 0.541 & 0.347 & 0.350 & 0.340 & 0.364 & 0.304 & 0.330 & 0.444 & 0.295 & 0.541 \\ 
 & Us3 & 0.338 & 0.343 & 0.347 & 0.335 & 0.359 & 0.302 & 0.325 & 0.442 & 0.294 & 0.529 \\ 
 & Us4 & 0.504 & 0.332 & 0.337 & 0.315 & 0.339 & 0.290 & 0.311 & 0.415 & 0.281 & 0.504 \\ 
 & Us5 & 0.298 & 0.321 & 0.322 & 0.302 & 0.324 & 0.275 & 0.295 & 0.403 & 0.268 & 0.448 \\ \hline
\end{tabular}